\documentclass[journal]{IEEEtran}
\usepackage{amsmath,amsfonts}
\usepackage{array}
\usepackage[caption=false,font=normalsize,labelfont=sf,textfont=sf]{subfig}
\usepackage{cite}
\usepackage{textcomp}
\usepackage{stfloats}
\usepackage{url}
\usepackage{verbatim}
\usepackage{graphicx}
\usepackage{booktabs}
\usepackage{bm}
\usepackage{adjustbox}
\usepackage{amssymb}
\usepackage{xcolor}
\usepackage{booktabs} 
\usepackage{adjustbox}
\usepackage{enumitem} 
\usepackage{algorithm}
\usepackage{algpseudocode}
\begin{document}

\title{Mask 2D-3D: Adaptive Dual-Masked Autoencoder Network for Image-to-Point Cloud Registration}

\author{
  Zhixin Cheng, Jiacheng Deng, Xiaotian Yin, Baoqun Yin, Richang Hong, Tianzhu Zhang%
  \thanks{Zhixin Cheng and Richang Hong is with the School of Computer Science and Technology, Hefei University of Technology, Hefei 230601, China. (e-mail: zxcheng@hfut.edu.cn; hongrc.hfut@gmail.com.)}
  \thanks{Jiacheng Deng, Baoqun Yin and Tianzhu Zhang are with the School of Information Science and Technology, University of Science and Technology of China, Hefei 230027, China (e-mail: chengzhixin@mail.ustc.edu.cn; dengjc@mail.ustc.edu.cn;  bqyin@ustc.edu.cn; tzzhang@ustc.edu.cn).}%
  \thanks{Xiaotian Yin is with the Institute of Advanced Technology, University of Science and Technology of China, Hefei 230027, China (e-mail: xiaotianyin@mail.ustc.edu.cn).}%
  \thanks{Corresponding author: Tianzhu Zhang and Richang Hong.}%
  \thanks{Manuscript received April 19, 2021; revised August 16, 2021.}
}

\markboth{Journal of \LaTeX\ Class Files,~Vol.~14, No.~8, August~2021}%
{Shell \MakeLowercase{\textit{et al.}}: A Sample Article Using IEEEtran.cls for IEEE Journals}


\maketitle

\begin{abstract}
Detection-free methods for image-to-point cloud registration are prone to erroneous correspondences caused by domain and modality discrepancies, limited sensitivity of feature extractors, and the presence of non-overlapping regions.
The Masked Autoencoder (MAE) has shown strong performance in visual representation for images and point clouds. It may be helpful to apply this approach to image-to-point cloud registration—a task that requires unified feature extraction and accurate cross-modal correspondences. Standard MAE's random masking may overlook key regions due to limited camera views, reducing registration effectiveness. To address this, we propose the Intermodal Dual-MAE Framework (ID-MAE) with a Similarity-based RL Masking Strategy (SRLM), which adaptively masks informative positions by leveraging cross-modal similarity and reinforcement learning, thus narrowing the modality gap.
{Our method enhances cross-modal representation learning by enforcing representation consistency during feature extraction, thereby enabling more reliable 2D–3D correspondence estimation.}
Experiments on RGB-D Scenes v2 and 7-Scenes benchmarks show that our method achieves state-of-the-art performance in image-to-point cloud registration.
\end{abstract}


\section{Introduction}
 \IEEEPARstart{I}mage-to-point cloud registration (I2P) aims to determine the rigid transformation from the point cloud to the camera coordinate system, which involves the cross-modal matching of image and point cloud, followed by a pose estimator to compute rotation and translation matrices. Such registration is essential for tasks like 3D reconstruction\cite{3dreconstruction,re1,triple,cheng30,cheng40}, SLAM \cite{slam,slam1,chensi,slam2,deng2}, and visual localization\cite{visuallocalization,vis1,cheng2025,cheng10,cheng20}.
 However, images are represented as regular, dense 2D grids, whereas point clouds are unordered, sparse, and irregular 3D points. The significant modality gap makes designing a model that effectively interacts between RGB and geometric modalities a challenging task.

\begin{figure}[!t]
\centering
\includegraphics[width=\columnwidth]{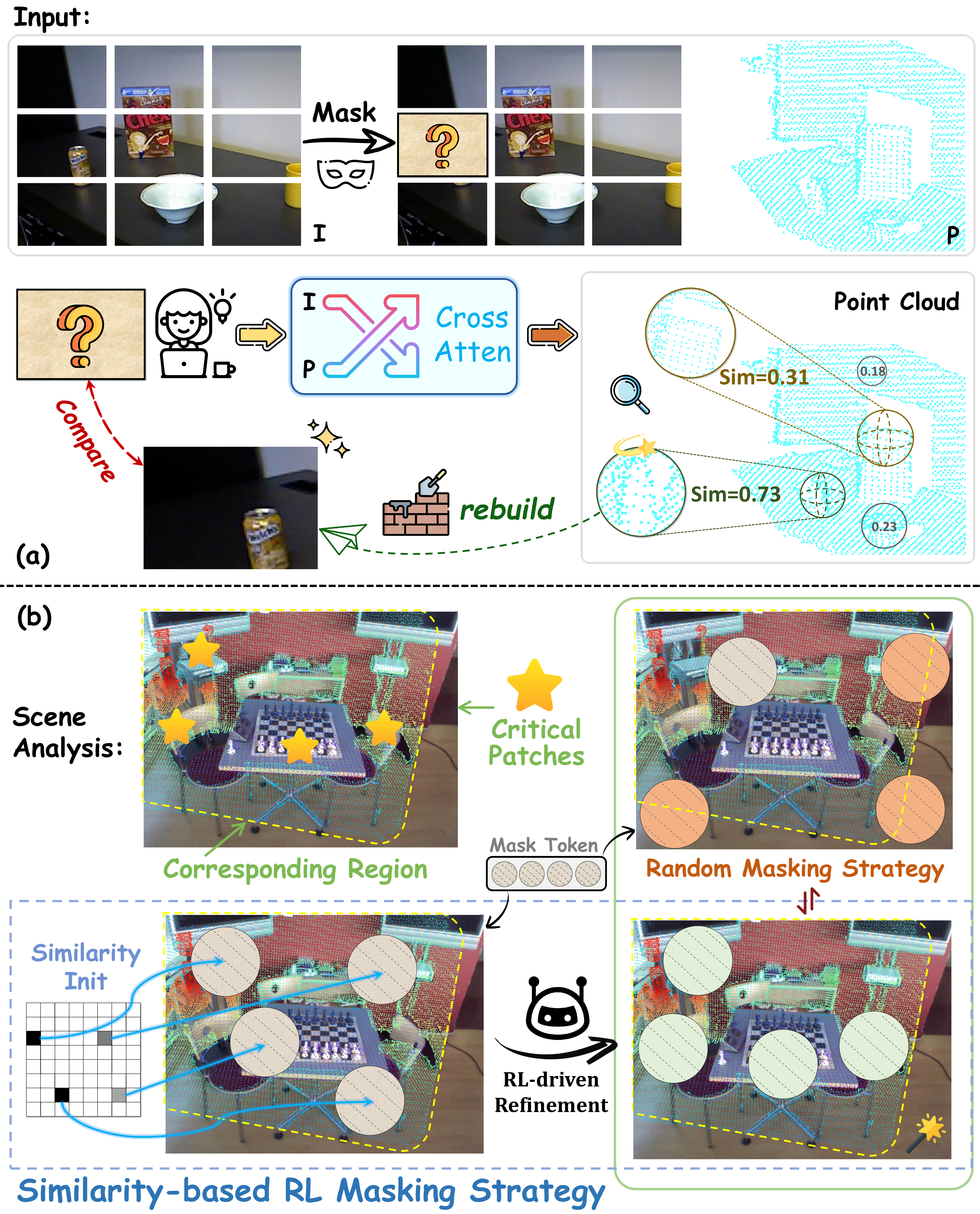} 
\caption{(a) Schematic of point cloud features aiding image feature reconstruction, where point cloud values denote attention-based correspondences.
(b) SRLM Strategy Diagram: Due to the presence of overlapping regions (within the yellow box) and high-value registration areas (star-shaped regions) in the scene, the orange blocks in the random masking strategy indicate ineffective masks. Therefore, we use similarity-based initialization to select patches (gray blocks) in the overlapping regions. Subsequently, we employ RL-driven refinement to seek more informative masking regions (green blocks). SRLM reduces redundant masking and focuses on critical matching regions.}
\label{fig1}
\end{figure}

Image-to-point cloud registration methods are typically classified into detect-then-match \cite{2d3dmatchnet, p2, deepi2p} and detection-free approaches \cite{cofii2p, corri2p}. Detect-then-match methods rely on independently detecting 2D and 3D keypoints and matching them via semantic features, but suffer from modality gaps and limited descriptor precision. Detection-free methods, such as 2D3D-MATR \cite{matr2d3d}, use a coarse-to-fine pipeline to establish patch-level matches and refine them to dense correspondences, improving inlier ratios via contextual cues and multi-scale receptive fields. However, image feature extraction relies on texture information, whereas point cloud feature extraction depends on structural information. As a result, the gaps between modalities and the limitations of feature extractors continue to impede further accuracy improvements.
Masked Autoencoders \cite{mae} have shown strong representation capability in both image and point cloud domains. However, despite some cross-modal attempts \cite{mae1,cheng1}, their potential for image-to-point cloud registration remains worth exploring. As a result, effectively leveraging MAE for cross-modal tasks remains challenging, since these tasks demand reconstructing features across different modalities while preserving precise correspondence alignment. Moreover, the commonly used random masking strategy may overlook critical regions or include irrelevant areas, limiting its ability to improve registration accuracy. Therefore, adapting both the MAE architecture and masking mechanism is essential to unlock its full potential for cross-modal alignment.

\begin{figure}[!t]
\centering
\includegraphics[width=\columnwidth]{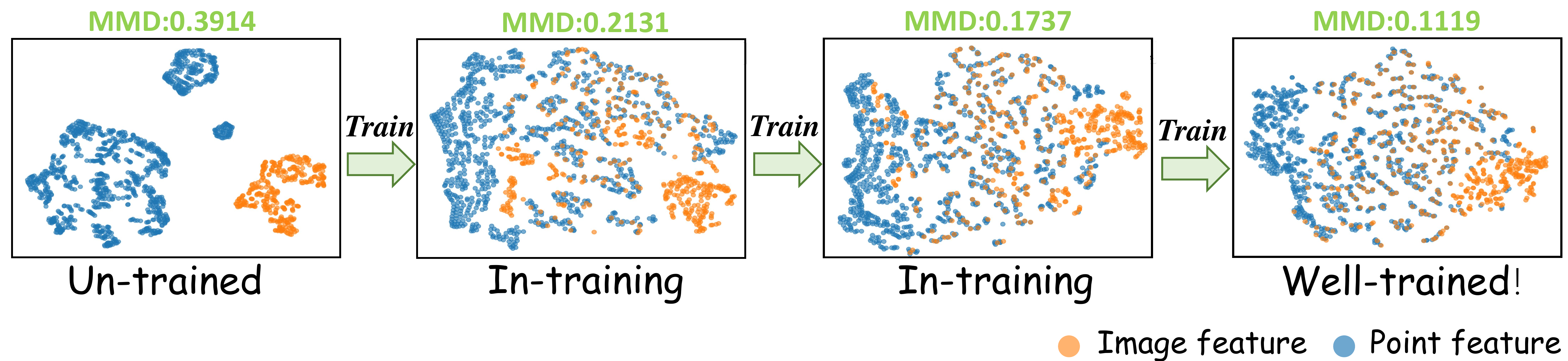} 
\caption{T-SNE visualization and MMD metrics (detailed in \ref{data}) showing improved alignment between image and point cloud features after training.}
\label{fig2}
\end{figure}

Based on the above discussions, we identify two key issues that need to be addressed to achieve accurate and robust image-to-point cloud registration.
\textbf{(1)} \textit{\textbf{How to design a unified framework that leverages MAE to enhance feature extraction across different modalities.} }
Although MAE has demonstrated remarkable improvements in feature extraction for single-modal tasks, designing a unified cross-modal MAE paradigm remains challenging. 
If we can enhance the feature extraction capability and further leverage cross-modal guidance for local structure reconstruction, it will greatly improve the ability to extract high-quality features and alleviate domain discrepancies (see in Fig.~\ref{fig2}), ultimately increasing the matching success rate.
As shown in Fig.~\ref{fig1}(a), the attention maps highlight the correspondences between point cloud and image features, with the point cloud features aiding in the reconstruction of the image features.
\textbf{(2)} \textit{\textbf{How to select masked regions to maximize the effectiveness of our design.}}
Random masking may be suboptimal, as it can mask irrelevant areas due to the camera's limited field of view, while missing key regions for registration. This not only increases computational cost but also degrades point cloud reconstruction, limiting the effectiveness of the MAE module. To address this, we propose a masking strategy focused on critical patches for registration in overlapping regions between the image and point cloud. As shown in Fig.~\ref{fig1}(b), our method reduces redundant masked areas (orange) and selects more informative regions (green).

To address the above challenges, we propose the Adaptive Dual-Masked Autoencoder Network for Image-to-Point Cloud Registration (M23D), featuring two core components: the Intermodal Dual-MAE Framework (ID-MAE) and the Similarity-based RL Masking Strategy (SRLM). ID-MAE introduces a unified cross-modal MAE with a bidirectional design, where point cloud features assist image reconstruction. Partial masking is applied to both modalities, and a low-resolution point cloud constrains the MAE-reconstructed output, enhancing feature quality. Image reconstruction is guided by both point cloud features and original image supervision, aligning modalities and establishing correspondences.
SRLM improves MAE’s effectiveness through a similarity-based initialization mechanism and RL-driven refinement. The masking strategy first targets overlapping regions by computing cross-modal similarity maps and selecting the top-$k$ most relevant areas for masking. Then, we apply reinforcement learning \cite{wang1,wang2,wang3} to make the selection of high-value regions differentiable, effectively focusing reconstruction on discriminative regions and improving registration accuracy.

{Unlike transformer-based matching methods such as 2D3D-MATR, which enhance correspondence estimation by refining interactions among pre-extracted features, our approach addresses cross-modality discrepancies earlier. Specifically, the proposed Intermodal Dual-MAE explicitly constrains feature learning via cross-modal reconstruction, encouraging image and point cloud representations to encode consistent geometric and semantic cues from the outset. Recent interaction-based methods have made notable progress in addressing the cross-modality challenge of image-to-point-cloud registration. In particular, Flow-I2P introduces an innovative information-geometric perspective, reformulating I2P registration as a manifold alignment problem and leveraging Beltrami flow to progressively refine cross-modal feature manifolds. This design effectively enhances feature interaction and improves generalization under limited training data. Our work is motivated by a complementary observation. Beyond refining feature interactions or manifold structures after feature extraction, the cross-modality gap can also be alleviated early in the representation learning process. Instead of explicitly modeling manifold evolution, we employ an intermodal dual-masked autoencoder. This method implicitly encodes cross-modal consistency through reconstruction, encouraging image and point cloud features to share aligned geometric semantics from the outset. In this sense, our approach focuses on learning more compatible representations before correspondence reasoning. It therefore complements existing interaction-based or manifold-alignment methods.}

In summary, our work can be summarized as follows:
\begin{itemize}
\item {We propose the Similarity Driven Dual-Masked Autoencoder Network (M23D), a unified architecture of the Intermodal Dual-MAE Framework (ID-MAE) with a novel Similarity-based RL Masking Strategy (SRLM) achieves strong accuracy and robustness. To our knowledge, this is the first MAE-based cross-modal design for image-to-point cloud registration.}
\item ID-MAE aligns modalities and establishes correspondences via a dual-MAE framework, while SDMS reduces modality gaps by selecting informative mask regions, improving efficiency and reconstruction quality.
\item Extensive experiments and ablations on RGB-D Scenes v2 and 7-Scenes demonstrate the effectiveness of our approach, setting a new state-of-the-art in image-to-point cloud registration.
\end{itemize}

\section{Related Work}
In this section, we briefly overview related works on image-to-point cloud registration, including stereo image registration, point cloud registration, and inter-modality registration.

\textbf{Stereo Image Registration.} Detector-based methods have long dominated stereo image registration. Prior to deep learning, key points were detected using handcrafted techniques like SIFT \cite{sift} and ORB \cite{orb}, which built 2D matches from local features. The advent of deep learning introduced neural network-based detection, transforming the field. SuperGlue \cite{superglue} was pioneering in using Transformers \cite{transformer} for image registration, greatly enhancing local feature matching. However, the challenge of detecting repeatable interest points in non-salient areas has led to the rise of detector-free methods. Approaches like LoFTR \cite{loftr} and Efficient LoFTR \cite{efficientloftr} use a coarse-to-fine pipeline with Transformers to efficiently estimate dense image matches through global receptive fields.

\begin{figure*}[!t]
    \centering
    \includegraphics[width=\textwidth]{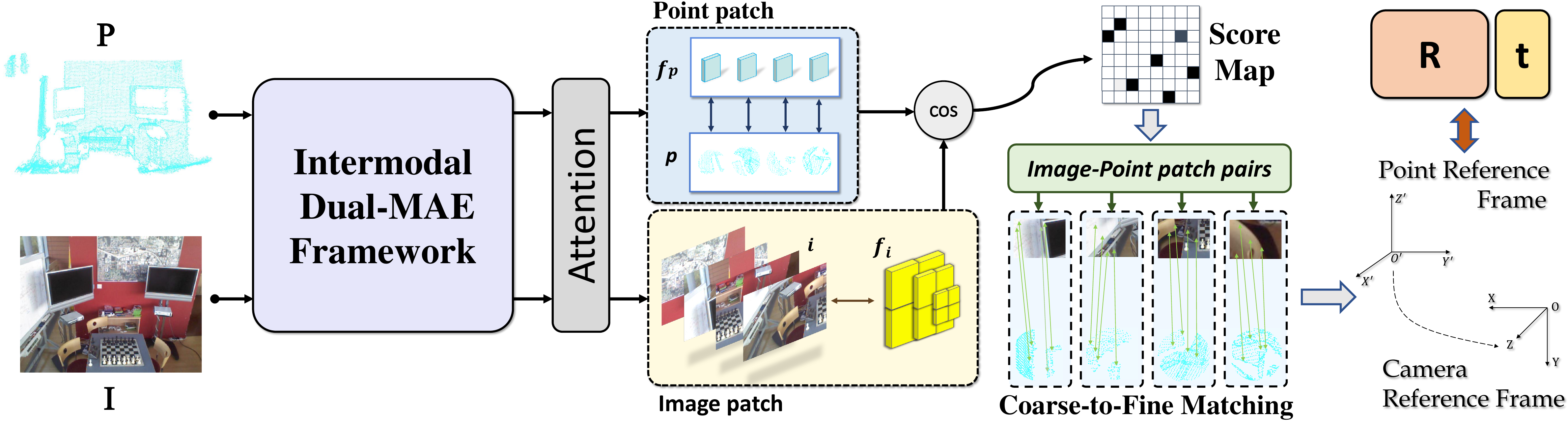} 
    \caption{Overall pipeline of M23D. We extract image and point cloud features through the intermodal dual-MAE framework, which are then processed via attention mechanism. The point cloud and image features generate a score map using cosine similarity and maximum operations, enabling coarse-level matching and refining fine-level correspondences. Finally, PnP + RANSAC is applied to estimate the rigid transformation.
}
\label{model1}
\end{figure*}

\textbf{Point Cloud Registration.} Point cloud registration has advanced from handcrafted descriptors like PPF \cite{ppf,ppf1} and FPFH \cite{fpfh} to deep learning-based approaches. CoFiNet \cite{cofinet} introduced detector-free registration with a coarse-to-fine strategy, and recent methods have replaced RANSAC \cite{ransac} with deep robust estimators for better speed and accuracy. GeoTransformer \cite{geotransformer} improves inlier ratios by integrating global information with the transformer and introduces a local-to-global method for RANSAC-free registration.

{\textbf{Cross-modal Masked Autoencoder Methods.}
Existing 2D–3D studies have explored cross-modal representation learning and MAE frameworks from different perspectives. PiMAE \cite{pimae} employs an interactive dual-branch MAE architecture, where image and point cloud tokens are reconstructed through complementary cross-modal masking based on geometric projection. Such a design enhances object-level semantic representations but relies on projection-based mask generation rather than adaptive selection of informative matching regions. I2P-MAE \cite{i2pmae} transfers knowledge from large-scale 2D pre-trained models through semantic-guided point masking and hierarchical point reconstruction, where visible point tokens are selected according to 2D semantic priors. Although effective for generic 3D representation learning, its masking strategy is determined by fixed semantic guidance rather than pair-specific cross-modal correspondence. CrossNet \cite{crossnet} aligns image and point cloud feature spaces via cross-modal contrastive learning, emphasizing global feature consistency and discriminative capability without adopting MAE-based masked reconstruction. Inter-MAE \cite{intermae} introduces image features to supervise point-cloud MAE pre-training through cross-modal contrastive learning while retaining conventional random masking for point reconstruction. Consequently, cross-modal interaction is mainly achieved through feature supervision instead of adaptive mask optimization. Joint-MAE \cite{jointmae} constructs a unified 2D–3D MAE by jointly encoding image and point cloud tokens and reconstructing both modalities with modality-specific decoders. It adopts random masking for both modalities and emphasizes generic multimodal pre-training, without explicitly considering overlap-aware masking or registration-oriented correspondence learning.

Overall, these methods demonstrate the effectiveness of MAE and cross-modal learning for 2D–3D representation modeling and provide valuable insights into cross-modal feature interaction. However, their architectures are primarily designed for generic representation learning or perception tasks, while their masking strategies are generally projection-based, random, or guided by fixed semantic priors. None of them explicitly optimize mask generation according to cross-modal overlap, correspondence reliability, or the registration objective. In contrast, our method introduces a dual-MAE framework tailored for image-to-point cloud registration, where image and point cloud features are jointly optimized through cross-modal reconstruction. Furthermore, the proposed Similarity-based RL Masking Strategy (SRLM) first initializes masks using bidirectional cross-modal similarity and then refines them via reinforcement learning, enabling adaptive selection of informative regions that directly benefit cross-modal correspondence learning and registration accuracy.

\textbf{Inter-modality Registration.} Inter-modal registration presents greater challenges compared to intra-modal registration because of the significant domain discrepancies involved. Traditional approaches typically employ a detect-then-match strategy. For instance, 2D3D-Matchnet \cite{2d3dmatchnet} uses SIFT \cite{sift} and ISS \cite{iss} to extract key points from images and point clouds, constructing patches around these key points. It then uses CNNs and PointNet \cite{pointnet} to extract features and build descriptors for matching. P2-Net \cite{p2} introduces a joint learning framework with a comprehensive reception mechanism, using a single forward pass to detect key locations and extract descriptors, enabling efficient matching through contrastive constraints. Unfortunately, the inefficiency of keypoint extraction in cross-modal scenarios has led to significant accuracy loss, prompting the emergence of detection-free methods. 2D3D-MATR \cite{matr2d3d} adopts a coarse-to-fine matching process, using a transformer-based approach to establish patch-level matches and then seeking fine-grained matches within them, followed by regressing rigid transformations using PNP+RANSAC \cite{pnp, ransac}. This detection-free method overcomes the challenge of obtaining repeatable key points and makes 2D-3D descriptors more consistent. Using transformers’ global receptive fields and multi-level methods significantly increases the inlier ratio of matches. B2-3Dnet \cite{b23d} further enhance cross-modal correspondence learning by leveraging covariance-guided feature alignment to improve the robustness and consistency of descriptors.
Based on this, CA-I2P \cite{cai2p} introduces channel adaptation and global optimal selection to better align cross-modal features and reduce redundant matches, achieving improved registration accuracy.
Flow-I2P \cite{flowi2p} improves image-to-point-cloud registration by using Beltrami flow for better manifold alignment and enhanced registration accuracy.
Our method advances detection-free approaches by introducing a dual MAE framework (ID-MAE) combined with a Similarity-based RL Masking Strategy (SRLM), where the dual MAE framework aligns modalities and establishes correspondences, while SRLM selectively masks high-value regions through similarity-based initialization and RL-driven refinement, making it a state-of-the-art solution for image-to-point cloud registration.

\section{Method}
\subsection{Overview} 
\begin{figure*}[!t]
\centering
\includegraphics[width=\textwidth]{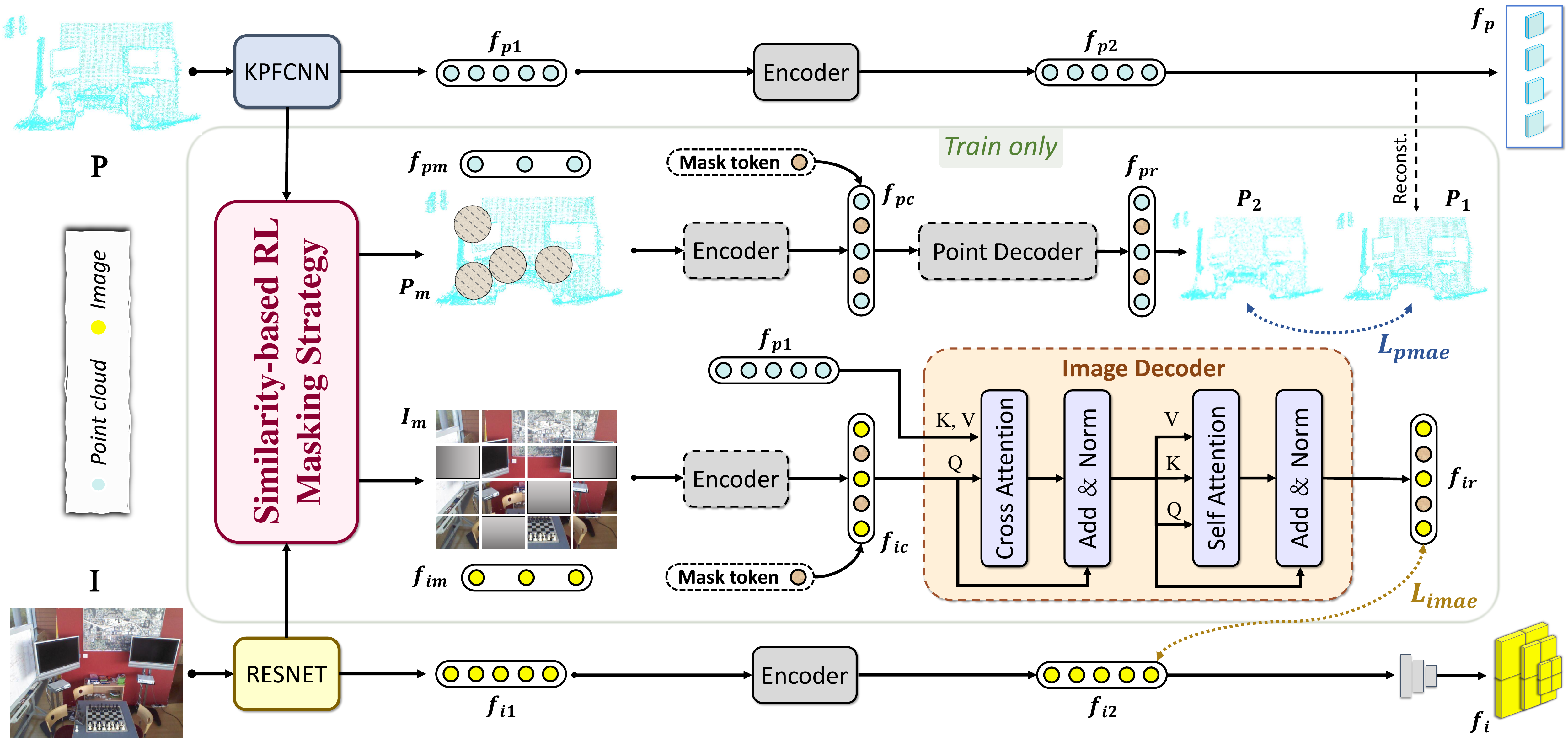} 
\caption{Overall pipeline of ID-MAE. After selecting the mask regions through SRLM, the point cloud MAE module constructs a loss constraint by reconstructing the point cloud from its low-resolution version. The image MAE module incorporates point cloud features to assist in reconstruction and is supervised by the original features. After post-processing, \(f_i\) and \(f_p\) are output for subsequent matching.}
\label{model2}
\end{figure*}

Given an image \(I \in \mathbb{R}^{H \times W \times 3}\) and a point cloud \(P \in \mathbb{R}^{N \times 3}\) from the same scene, the objective of the registration between images and point clouds is to determine the rigid transformation \([R, t]\) from the point cloud coordinate system to the camera coordinate system. Here, \(W\) and \(H\) are the image's width and height, while \(N\) is the number of points. The transformation consists of a 3D rotation \(R \in \mathit{SO}(3)\) and a 3D translation vector \(t \in \mathbb{R}^3\).

Our method, M2-3D, depicted in Fig.~\ref{model1}, adopt a dual MAE-enhanced detection-free registration paradigm, constructing the modality-aligned and corresponded image and point cloud features through the intermodal dual-MAE framework. After passing through a series of attention modules, similarity is calculated to generate a score map, enabling patch-level matches. High-resolution images and point cloud features are then used to refine dense correspondences from the patch matches. Finally, the PnP+RANSAC \cite{pnp} algorithm effectively regresses the rigid transformation.

\subsection{Intermodal Dual-MAE Framework}
We utilize ResNet \cite{resnet} with FPN \cite{fpn} and KPFCNN \cite{kpfcnn} to extract features from images and point clouds, respectively. The 2D features and 3D features are downsampled at the lowest resolution to obtain \(f_{i1} \in \mathbb{R}^{(h \times w) \times c}\) and \(f_{p1} \in \mathbb{R}^{n \times c}\). For the extracted features at the lowest resolution \(f_i\) and \(f_p\), positional encoding is applied to enhance the features.
To prepare for the subsequent MAE, we do not adopt the conventional pretraining mode of MAE, as pretraining significantly increases computational overhead and is often constrained in certain scenarios. Instead, we aim to enhance the feature extraction capability of the network encoder through unsupervised MAE during normal training. This approach facilitates modality aggregation and establishes correspondences between images and point clouds. We can observe the overall pipeline of Intermodal Dual-MAE Framework (ID-MAE) from Fig.~\ref{model2}.

\subsubsection{\bfseries Dual-modal MAE Enhancement}
Point MAE module adopts a self-supervised process, where the point cloud feature \(f_{p1}\) is masked based on the Similarity-based RL Masking Strategy (SRLM) (we will elaborate on this later), resulting in \(f_{pm} \in \mathbb{R}^{(n-m_p) \times c}\), where \(m_p\) denotes the number of masked points. For the encoder, we employ local self-attention instead of global self-attention, to construct a Transformer-based encoder. We fill the masked positions in \(f_{pm}\) with learnable mask tokens of dimension \(\mathbb{R}^{1\times c}\), completing the feature to obtain \(f_{pc} \in \mathbb{R}^{n \times c}\) with the exact dimensions as \(f_{p1}\). After passing through the decoder, the mask tokens learn the corresponding point cloud features, completing the point cloud reconstruction process.
We utilize standard point cloud cross-attention blocks in the point cloud decoder, along with self-attention blocks. The reconstructed point cloud features \(f_{p1} \in \mathbb{R}^{n \times c}\) are finally mapped to the point cloud \(P_2 \in \mathbb{R}^{n \times 3}\), while the features \(f_{p2}\) obtained from the encoder are decoded to \(P_1 \in \mathbb{R}^{n \times 3}\). The two point clouds \(P_1\) and \(P_2\) are supervised by an $L_2$ loss. Here, \(n\) denotes the number of points in the reconstructed point cloud, and \(m\) indexes the \(m\)-th point.
\begin{equation}
\mathcal{L}_{\text{pmae}} = \frac{1}{M} \sum_{m=1}^M \| P_{1,m} - P_{2,m} \|_2^2 ,
\end{equation}
after filtering out some excessively small point cloud patches, we output the point cloud feature \(f_p\) for subsequent matching. {As illustrated in Fig.~4, we adopt a KPFCNN backbone that naturally produces a hierarchical multi-scale point pyramid from the input point cloud $P$. The point MAE operates on the coarsest-scale level of this pyramid as its reconstruction target, while finer-scale point features are preserved for subsequent matching and registration. Therefore, the effective downsampling ratio of the low-resolution point set is implicitly determined by the backbone’s subsampling configuration, and can be reproduced by reporting the actual point counts at different hierarchy levels.
Our masking strategy further selects informative regions on the coarsest-level point tokens, producing masked subsets $P_m$ for the training-only MAE branch. At the dataset level, a random subsampling is optionally applied only to cap the maximum number of raw input points for efficiency; this preprocessing neither defines the low-resolution point cloud used in ID-MAE nor affects the hierarchical structure. Importantly, the low-resolution point cloud is solely used as a reconstruction target during training and is removed at inference, thus it does not participate in similarity computation, masking selection, or pose estimation. This design naturally aligns with the coarse-to-fine registration paradigm, where global structural consistency is first captured at coarse resolution and progressively refined at finer scales.
}

For point clouds, we reconstruct them for supervision due to their sparsity, while for images, we directly supervise the features.
And since the point cloud perspective contains more information, we aim to introduce cross-modal information to assist in the reconstruction of the image.

\subsubsection{\bfseries Cross-Modal Guided Reconstruction}
Typically, MAE encoders benefit from learning a generalized encoder capable of capturing high-dimensional data representations of both images and point clouds. Due to the differences between the two modalities, a dedicated decoder is required to decode the high-level latent data across two different modalities. 
The encoder part of image MAE module is similar to that of the point cloud. The image feature \(f_{i1}\) undergoes SRLM, resulting in \(f_{im} \in \mathbb{R}^{(h \times w - m_i) \times c}\),  where \(m_i\) denotes the number of masked patches. 

We leverage point cloud features to restore the masked regions in the image, facilitating the identification of corresponding structures between the two modalities and enhancing semantic alignment. A learnable mask token of dimension \(\mathbb{R}^{1 \times c }\) is introduced and duplicated to fill the missing positions in the masked image features, yielding a complete feature map \(f_{ic} \in \mathbb{R}^{(h \times w) \times c}\). The feature \(f_{ic}\) is used as the query (\(Q\)), while the point cloud representation \(f_{p1}\) serves as both the key (\(K\)) and value (\(V\)) in the cross-attention mechanism to reconstruct image features.

These components are jointly fed into the decoder to model inter-modal dependencies and produce the updated and restored feature \(f_{ir} \in \mathbb{R}^{(h \times w) \times c}\). This decoding process enhances alignment and improves reconstruction accuracy. The restored feature \(f_{ir}\) is obtained by aggregating relevant point cloud features through a weighted summation based on feature similarity, as defined below:
\begin{equation}
f_{ir} = \sum_{b \in \mathcal{N}_{f_{p1}}(a)} \frac{e^{s_{ab}}}{\sum_{l \in \mathcal{N}_{f_{p1}}(a)} e^{s_{al}}} f_{ic}^b,
\label{eq:feature_reconstruction}
\end{equation}
where \(\mathcal{N}_{f_{p1}}(a)\) denotes the neighborhood of the \(a\)-th feature in \(f_{p1}\) associated with \(f_{ic}\). The similarity \(s_{ab}\) between \(f_{p1}^a\) and \(f_{ic}^b\) is computed as:
\begin{equation}
s_{ab} = {f_{p1}^a}^\top W f_{ic}^b,
\label{eq:similarity_score}
\end{equation}
where \(W\) is a learnable parameter matrix. This formula calculates the similarity between \(f_{ic}\) and \(f_{p1}\) features, applies a weighted summation over the neighboring image features \(f_{p1}\), and generates the reconstructed target feature \(f_{ir}\). This process effectively fuses the point cloud and image features, providing the necessary support for subsequent decoding. The \(f_{i1}\) feature is passed through the encoder to obtain \(f_{i2} \in \mathbb{R}^{(h \times w) \times c}\), which is constrained during the reconstruction process. \(N\) represents the number of patches extracted from the image, corresponding to the total number of feature tokens in the first dimension of \(f_{i2}\). The \(n\)-th patch refers to a specific patch among these \(N\) patches.
\begin{equation}
\mathcal{L}_{\text{imae}} = \frac{1}{N} \sum_{n=1}^N \left(f_{i2,n} - f_{ir,n}\right)^2, 
\end{equation}
a lightweight three-stage CNN \cite{lightweightcnn} is used to extract image features at three different scales, capturing multi-level spatial information. The resulting features form a feature pyramid denoted as \(f_i\), which serves as input for subsequent process.

\textbf{Discussion.}
\textit{Why do we choose to reconstruct image features using point cloud features rather than the other way around?} The reason lies in the fact that point clouds and images do not correspond exactly. Compared to images, point clouds provide a broader scene view and richer information. Thus, reconstructing point cloud features from images may suffer from information loss, degrading reconstruction quality and constraints. Moreover, restricting point cloud features to the image range requires ground truth $R, t$, which we aim to predict, making this approach unreasonable.

\begin{figure}[t]
\centering
\includegraphics[width=\columnwidth]{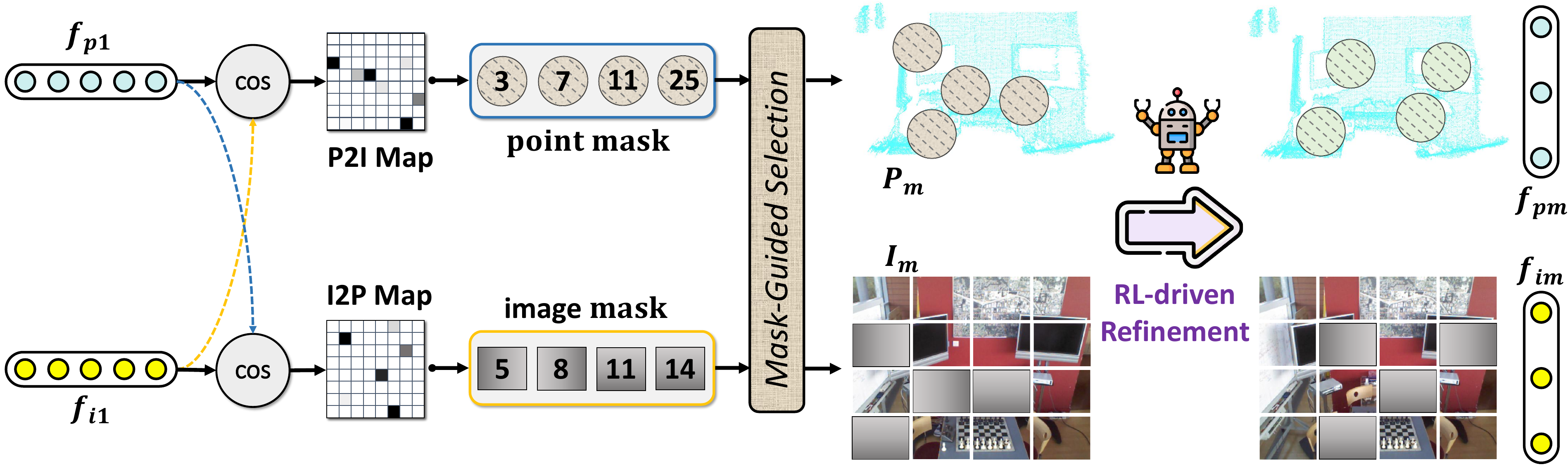}
\caption{Overall pipeline of SRLM, where similarity-based masking selection is used for prior initialization, followed by RL-driven refinement.}
\label{model3}
\end{figure}

\subsection{Similarity-based RL Masking Strategy}
\label{SRLM}

In image-to-point-cloud registration, due to the existence of non-overlapping regions in the image and point cloud views, when image or point cloud patches fall within these areas, it can lead to completely incorrect matches, making these computations meaningless. Additionally, since the approach follows a coarse-to-fine paradigm, this can result in large misalignments during fine-grained matching, which is undesirable. Therefore, we aim to increase the likelihood of selecting regions within the overlapping areas through similarity-based selection. However, due to domain differences in image and point cloud features, areas with high similarity cannot be directly considered as critical matching regions. Since the task of selecting key regions is inherently non-differentiable, we aim to address this challenge by employing a reinforcement learning strategy, allowing us to effectively solve the problem.
Thus to improve cross-modal learning, we propose the Similarity-based RL Masking Strategy (SRLM), which masks regions of critical patches. SRLM comprises a Similarity-based Masking Selection (SMS) and RL-driven Refinement (RLR).
The overall process of SDMS is presented in Fig. 5.

\subsubsection{\bfseries Similarity-based Masking Selection}
The camera's limited field of view may miss discriminative areas for registration. Expanding the masked regions increases computational costs and degrades point cloud reconstruction. Observing the matching process, we find that correspondences may rely on high similarity between image and point cloud features, which should be masked as initial state to enhance the model's perception.
We compute the similarity maps: the Image-to-Point Map (I2P Map) and the Point-to-Image Map (P2I Map). By setting similarity thresholds and performing top-$k$ selection, we determine the positions to mask for both images and point clouds, denoted as image mask and point mask, respectively. Using these guided positions, we apply Mask-Guided Selection, masking the corresponding areas to obtain $I_m$ and $P_m$.
Through the above design, we roughly determine the initial positions of the image-point cloud mask blocks, which will be used as the initial state input in the subsequent reinforcement learning process.
For features $f_{im} \in \mathbb{R}^{(h \times w - m_i) \times c}$ and $f_{pm} \in \mathbb{R}^{(n - m_p) \times c}$, $m_i$ and $m_p$ represent the number of masked regions in the image and point cloud, respectively.

{While similarity evaluation is indeed widely used in registration tasks, it is typically applied at the correspondence or matching stage. In contrast, our similarity-aware masking strategy leverages such cues during representation learning, guiding the MAE to focus on informative cross-modal regions before correspondence estimation.
This shift in usage is crucial: rather than directly establishing correspondences, similarity is used to prioritize which regions should be preserved or reconstructed during masked modeling, enabling the MAE to learn more correspondence-aware representations. However, due to repetitive object textures in the scene and domain gaps in the initialization between images and point clouds, a similarity-only selection strategy may not be sufficient to fully unleash the potential of our bidirectional MAE architecture.
}

{\subsubsection{\bfseries RL-driven Refinement}

To enhance the accuracy of image-to-point cloud matching, we introduce a reinforcement learning (RL)-based strategy optimization to refine the selection of masked regions (as shown in Algorithm 1). Instead of relying on fixed or heuristic masking schemes, the proposed RL framework adaptively adjusts which regions to mask or preserve during training, enabling the model to focus on more informative and reliable cross-modal areas.
The RL component is designed as an auxiliary optimization mechanism that operates alongside the MAE framework, rather than a standalone decision-making module. By observing the current similarity-aware state, the policy network dynamically refines the selection of masked regions, thereby reducing masking randomness and stabilizing MAE optimization. A task-aligned reward function derived from the matching loss guides policy updates, ensuring that the learned masking strategy consistently supports the main registration objective.
Through this design, the RL-based masking strategy facilitates more reliable feature learning for image-to-point cloud matching while avoiding additional inference overhead, as it is only applied during training.


The initial state in the RL process is determined by the similarity obtained from SMS. The reward function plays a critical role in this process, as it guides the optimization of the policy. In our approach, the reward is computed based on \( L_i \) (described in Equation~\ref{Li} in \ref{inference}), where the circle loss effectively constrains overall matching accuracy, making it an ideal candidate for the reward. Specifically, since better matching performance results in smaller values for \( L_i \), the reward increases as matching accuracy improves. Thus, the reward is defined as:
\begin{equation}
    R = \frac{1}{L_i + \delta},
\end{equation}
where \( \delta \) is a small constant to prevent division by zero.

The probabilities of selecting image and point cloud masks, denoted as \( Prob_i \) and \( Prob_p \), are computed based on the output of the policy network for both regions. These probabilities reflect the network's confidence in selecting each region and are calculated as follows:
\begin{equation}
    Prob_i = \log(\pi_\theta(a_{\text{imask}} | s_{\text{image}})),
\end{equation}
\begin{equation}
    Prob_p = \log(\pi_\theta(a_{\text{pmask}} | s_{\text{point}})),
\end{equation}
where \( \pi_\theta(a_{\text{imask}} | s_{\text{image}}) \) and \( \pi_\theta(a_{\text{pmask}} | s_{\text{point}}) \) represent the probabilities output by the policy network for the image and point cloud mask selections, respectively.

The overall log-probability, \( \text{logprob} \), is computed by averaging the log-probabilities of both modalities as follows:
\begin{equation}
    \footnotesize
    \log^{prob} = \frac{1}{2} \log\left[ 1 + \left(\frac{1}{h \cdot w} \sum_{i=1}^{h \cdot w} Prob_i \right) \cdot \left(\frac{1}{n} \sum_{j=1}^{n} Prob_p \right) \right],
\end{equation}
where \( h \cdot w \) and \( n \) represent the number of elements in the image and point cloud regions, respectively. This averaging process ensures that both image and point cloud regions contribute equally to the overall decision-making process by combining their confidence levels.

In RL, the action corresponds to selecting the masked regions based on the current state. In our approach, actions involve determining which regions should be masked and which should remain visible. The action space consists of all possible combinations of masked regions, and the policy network learns to select the optimal regions by maximizing the reward. The network computes the log-probability of the selected action, reflecting its confidence in the chosen masked regions. Specifically, the gradient of the policy is estimated as:
\begin{equation}
    \nabla \theta J(\theta) = \mathbb{E}_t \left[ \nabla \log \pi_\theta(a_t | s_t) \cdot R_t \right],
\end{equation}
where \( \pi_\theta(a_t | s_t) \) represents the probability of selecting action \( a_t \) given state \( s_t \), parameterized by \( \theta \). The log-probability \( \nabla \log \pi_\theta(a_t | s_t) \) is used to compute the gradient for policy update, while \( R_t \) is the reward associated with the action taken at time \( t \), guiding the policy towards optimal actions.

The reinforcement loss \( L_t \) is then computed as:
\begin{equation}
    L_t = - log^{prob} \times R.
\end{equation}

This policy loss is added to the main loss, steering the optimization of the network parameters. By maximizing the reward and minimizing the total loss, the policy network gradually learns to select the best masked regions, ultimately improving the matching accuracy between the image and point cloud.
This design helps stabilize the MAE training process by reducing randomness in mask selection and alleviating sensitivity to hyperparameters. The resulting adaptive masking strategy facilitates more reliable image-to-point cloud matching and improves robustness in practice.

\begin{algorithm}[H]
\footnotesize
\caption{RL-driven Refinement for Mask Region Selection}
\label{alg:rl_refinement}

\setlength{\abovedisplayskip}{2pt}
\setlength{\belowdisplayskip}{2pt}
\setlength{\abovedisplayshortskip}{1pt}
\setlength{\belowdisplayshortskip}{1pt}

\textbf{Input:}
Image feature $\mathbf{F}_I$, point-cloud feature $\mathbf{F}_P$,
SMS, policy network $\pi_{\theta}$, MAE network $\mathcal{M}$,
matching network $\mathcal{G}$, and constant $\delta$.

\textbf{Output:}
Image mask $a_{\mathrm{imask}}$ and point-cloud mask
$a_{\mathrm{pmask}}$.

\textbf{Similarity-aware initialization:}\\[-2pt]
\[
(s_{\mathrm{image}},s_{\mathrm{point}})
\leftarrow
\operatorname{SMS}(\mathbf{F}_I,\mathbf{F}_P).
\]

\textbf{Policy-based mask refinement:}
\begin{itemize}
\setlength{\itemsep}{1pt}
\setlength{\parsep}{0pt}
\setlength{\parskip}{0pt}
\setlength{\topsep}{2pt}

\item Predict mask-selection distributions:
\[
\pi_{\theta}(a_{\mathrm{imask}}\mid s_{\mathrm{image}}),
\qquad
\pi_{\theta}(a_{\mathrm{pmask}}\mid s_{\mathrm{point}}).
\]

\item Sample masking actions:
\[
a_{\mathrm{imask}}
\sim\pi_{\theta}(\cdot\mid s_{\mathrm{image}}),
\qquad
a_{\mathrm{pmask}}
\sim\pi_{\theta}(\cdot\mid s_{\mathrm{point}}).
\]

\item Apply the masks:
\[
\widetilde{\mathbf{F}}_I
\leftarrow
\operatorname{Mask}(\mathbf{F}_I,a_{\mathrm{imask}}),
\qquad
\widetilde{\mathbf{F}}_P
\leftarrow
\operatorname{Mask}(\mathbf{F}_P,a_{\mathrm{pmask}}).
\]
\end{itemize}

\textbf{Reconstruction and matching:}
\begin{itemize}
\setlength{\itemsep}{1pt}
\setlength{\parsep}{0pt}
\setlength{\parskip}{0pt}
\setlength{\topsep}{2pt}

\item Reconstruct the masked features:
\[
(\mathbf{F}_I^{r},\mathbf{F}_P^{r})
\leftarrow
\mathcal{M}(\widetilde{\mathbf{F}}_I,\widetilde{\mathbf{F}}_P).
\]

\item Estimate correspondences and matching loss:
\[
\mathcal{C}
\leftarrow
\mathcal{G}(\mathbf{F}_I^{r},\mathbf{F}_P^{r}),
\qquad
L_i\leftarrow\operatorname{CircleLoss}(\mathcal{C}).
\]
\end{itemize}

\textbf{Reward:}\\[-2pt]
\[
R\leftarrow\frac{1}{L_i+\delta}.
\]

\textbf{Policy probability:}
\begin{itemize}
\setlength{\itemsep}{1pt}
\setlength{\parsep}{0pt}
\setlength{\parskip}{0pt}
\setlength{\topsep}{2pt}

\item Compute modal log-probabilities:
\[
\begin{aligned}
Prob_i
&\leftarrow
\log \pi_{\theta}
\left(
a_{\mathrm{imask}}
\mid s_{\mathrm{image}}
\right),\\
Prob_p
&\leftarrow
\log \pi_{\theta}
\left(
a_{\mathrm{pmask}}
\mid s_{\mathrm{point}}
\right).
\end{aligned}
\]

\item Aggregate the two modalities:
\[
\log^{prob}
\leftarrow
\frac{1}{2}
\log\!\left[
1+
\left(
\frac{1}{hw}\sum_{i=1}^{hw}Prob_i
\right)
\left(
\frac{1}{n}\sum_{j=1}^{n}Prob_p
\right)
\right].
\]
\end{itemize}

\textbf{Policy optimization:}
\begin{itemize}
\setlength{\itemsep}{1pt}
\setlength{\parsep}{0pt}
\setlength{\parskip}{0pt}
\setlength{\topsep}{2pt}

\item Compute the policy and total losses:
\[
L_t\leftarrow-\log^{prob}R,
\qquad
L_{\mathrm{total}}
\leftarrow
L_{\mathrm{main}}+\lambda_tL_t.
\]

\item Update the policy and main-network parameters:
\[
\theta
\leftarrow
\theta-\eta_{\theta}\nabla_{\theta}L_t,
\qquad
\phi
\leftarrow
\phi-\eta_{\phi}\nabla_{\phi}L_{\mathrm{total}}.
\]
\end{itemize}

\textbf{Return:}
$a_{\mathrm{imask}}$ and $a_{\mathrm{pmask}}$.

\end{algorithm}

\textbf{Training stability discussion.}
Although the RL module is tightly coupled with the registration task, several design choices are adopted to ensure stable optimization.
First, the RL policy is not activated from the beginning of training. As shown in our implementation, the network is first trained with MAE-based reconstruction losses during a warm-up phase (set for the first 20 epochs), allowing cross-modal representations to stabilize before policy learning is introduced. This significantly reduces the variance and sensitivity of subsequent policy updates.
Second, the reward signal is dense and continuous, defined as the inverse of the main task loss at each iteration. Unlike sparse or delayed rewards, this design provides immediate and consistent feedback to the policy, effectively mitigating reward sparsity.
Third, the policy loss is incorporated as a lightweight regularization term with a small weighting factor, ensuring that the primary optimization objective remains dominated by the registration loss. As a result, the RL component softly biases the masking strategy rather than aggressively steering the feature-learning process, reducing the risk of co-adaptation or training collapse.
To further improve stability, the policy employs factorized Bernoulli sampling at the token level, and the log-probabilities of mask selections are averaged across multiple token-wise decisions. This design significantly reduces policy gradient variance compared to combinatorial masking actions.
Finally, the policy state is initialized and constrained by similarity-based priors, limiting exploration to informative regions and further mitigating co-adaptation with the main network. As a result, the RL component serves as a lightweight refinement mechanism rather than a dominant optimization driver.}

\subsection{Model Training \& Inference}
\label{inference}
Our training consists of two stages: first, we train the point MAE module, followed by the image MAE module. As image feature reconstruction depends on point cloud features, we prioritize training the point MAE module to ensure high-quality feature extraction. Our ID-MAE is employed solely during training and omitted at inference.

To obtain the total loss, first let us examine the loss functions for the coarse and fine matching networks. Both \(\mathcal{L}_{\text{coarse}}\) and \(\mathcal{L}_{\text{fine}}\) use the general circle loss \cite{circleloss,circleloss1}. Given an anchor descriptor \(d_i\), the descriptors of its positive and negative pairs are \(\mathcal{D}^P_i\) and \(\mathcal{D}^N_i\), respectively:
\begin{equation}
\footnotesize
\mathcal{L}_i = \frac{1}{\gamma} \log\left[ 1 + \Bigl( \sum_{d^j \in \mathcal{D}^P_i} e^{\beta^{i,j}_p(d^j_i - \Delta_p)} \Bigr) \cdot \Bigl( \sum_{d^k \in \mathcal{D}^N_i} e^{\beta^{i,k}_n(\Delta_n - d^k_i)} \Bigr) \right],
\label{Li}
\end{equation}

where \(d^j_i\) is the \(L_2\) feature distance, \(\beta^{i,j}_p = \gamma \lambda^{i,j}_p(d^j_i - \Delta_p)\) and \(\beta^{i,k}_n = \gamma \lambda^{i,k}_n(\Delta_n - d^k_i)\) are the individual weights for the positive and negative pairs, with \(\lambda^{i,j}_p\) and \(\lambda^{i,k}_n\) as scaling factors \cite{overlap}.

Combining the discussions above, the total loss is composed of three key components: the MAE loss for image and point cloud ($\mathcal{L}_{\text{imae}}$, $\mathcal{L}_{\text{pmae}}$), the aggregation loss $\mathcal{L}_t$ from the SRLM module, and the matching loss $\mathcal{L}_d$ (including \(\mathcal{L}_{\text{coarse}}\) and \(\mathcal{L}_{\text{fine}}\)) from the matching process. The total loss is computed as:
\begin{equation}
\mathcal{L}_{\text{total}} = \gamma_1 \mathcal{L}_{\text{imae}} + \gamma_2 \mathcal{L}_{\text{pmae}} + \gamma_3 \mathcal{L}_t + \gamma_4 \mathcal{L}_d,
\label{eq:total_loss}
\end{equation}
where $\gamma_i$ are hyperparameters balancing the contribution of different loss terms.

\section{Experiments}

\subsection{Datasets and Implementation Details}
\label{data}
Based on the 2D3D-MATR benchmark, we conducted extensive experiments and ablation studies on two challenging benchmarks: RGB-D Scenes v2 \cite{rgbdv2} and 7Scenes \cite{7scenes}.

\textbf{Dataset. } \textit{RGB-D Scenes Dataset v2} consists of 14 scenes containing furniture (chair, coffee table, sofa, table) and a subset of the objects in the RGB-D Object Dataset (bowls, caps, cereal boxes, coffee mugs, and soda cans). For each scene, we create point cloud fragments from every 25 consecutive depth frames and sample one RGB image per 25 frames. We select image-point-cloud pairs with an overlap ratio of at least 30\%. Scenes 1-8 are used for training, 9-10 for validation, and 11-14 for testing, resulting in 1,748 training pairs, 236 validation pairs, and 497 testing pairs.

The \textit{7-Scenes dataset} is a collection of tracked RGB-D camera frames. All 7 indoor scenes were recorded from a handheld Kinect RGB-D camera at 640×480 resolution. We select image to point-cloud pairs from each scene with at least 50\% overlap, adhering to the official sequence split for training, validation, and testing. This results in 4,048 training pairs, 1,011 validation pairs, and 2,304 testing pairs.

The \textit{KITTI Odometry} dataset contains 22 image and point cloud sequences, with 11 sequences providing ground-truth calibration. 
We use sequences 00--08 for training and 09--10 for testing. 
To simulate mis-registration, a 2D translation within $\pm 10$\,m and an unconstrained rotation around the up-axis are applied. 
Images are downsampled to $160 \times 512$ and point clouds to 40,960 points.

\textbf{T-SNE \& MMD.} t-SNE \cite{tsne} is a dimensionality reduction technique that projects high-dimensional data into 2D or 3D while preserving local structure, making it useful for visualizing feature distributions. MMD is a non-parametric metric that measures the difference between two distributions in a kernel-based feature space. Both are commonly used in tasks such as domain adaptation and cross-modal representation analysis.

\textbf{Implementation Details.} We use an NVIDIA Geforce RTX 3090 GPU for training. The entire pipeline is implemented using PyTorch.
In the image decoder, the cross-attention module is applied once, while the self-attention module is applied three times. The threshold for the I2P map is set to 0.5, with $k_1 = 35$ in the top-$k$ selection. Similarly, the threshold for the P2I map is set to 0.5, with $k_2 = 15$ in the top-$k$ selection. For loss $\gamma_1 = \gamma_2 = \gamma_3 = \gamma_4 = 1$. 



\begin{table}[!t]
\caption{Evaluation results on RGB-D Scenes v2. \textcolor{teal}{Teal} numbers highlight the best, the second best are \textbf{Boldfaced} and the baseline are \underline{underlined}.}
\centering
\renewcommand{\arraystretch}{0.95}
\setlength{\tabcolsep}{1.68mm} 
\resizebox{\columnwidth}{!}{ 
\begin{tabular}{lccccc}
\toprule\toprule
\multicolumn{1}{l|}{Model} & Scene.11 & Scene.12 & Scene.13 & Scene.14 & Mean \\ \midrule
\multicolumn{1}{l|}{Mean depth (m)} & 1.74 & 1.66 & 1.18 & 1.39 & 1.49 \\ \midrule
\multicolumn{6}{c}{\textit{Inlier Ratio} ↑} \\ \midrule
\multicolumn{1}{l|}{FCGF-2D3D \cite{fcgf2d3d}} & 6.8 & 8.5 & 11.8 & 5.4 & 8.1 \\
\multicolumn{1}{l|}{P2-Net \cite{p2}} & 9.7 & 12.8 & 17.0 & 9.3 & 12.2 \\
\multicolumn{1}{l|}{Predator-2D3D \cite{predator2d3d}} & 17.7 & 19.4 & 17.2 & 8.4 & 15.7 \\
\multicolumn{1}{l|}{2D3D-MATR \cite{matr2d3d}} & \underline{32.8} & \underline{34.4} & \underline{39.2} & \underline{23.3} & \underline{32.4} \\
\multicolumn{1}{l|}{FreeReg \cite{freereg}} & 36.6 & 34.5 & 34.2 & 18.2 & 30.9 \\
\multicolumn{1}{l|}{B2-3Dnet \cite{b23d}} & 36.4 & 32.7 & \textbf{43.8} & 27.4 & 35.1 \\ 
\multicolumn{1}{l|}{CA-I2P \cite{cai2p}} & 38.6 & 40.6 & 38.9 & 24.0 & 35.5 \\
\multicolumn{1}{l|}{Flow-I2P \cite{flowi2p}} & \textcolor{teal}{\textbf{49.6}} & \textbf{44.0} & 36.5 & \textbf{30.4} & \textbf{40.1} \\
\multicolumn{1}{l|}{M23D (ours)} & 
\textbf{48.1} & 
\textcolor{teal}{\textbf{48.4}} & 
\textcolor{teal}{\textbf{46.3}} & 
\textcolor{teal}{\textbf{31.4}} & 
\textcolor{teal}{\textbf{43.5}} \\ \midrule
\multicolumn{6}{c}{\textit{Feature Matching Recall} ↑} \\ \midrule
\multicolumn{1}{l|}{FCGF-2D3D \cite{fcgf2d3d}} & 11.1 & 30.4 & 51.5 & 15.5 & 27.1 \\
\multicolumn{1}{l|}{P2-Net \cite{p2}} & 48.6 & 65.7 & 82.5 & 41.6 & 59.6 \\
\multicolumn{1}{l|}{Predator-2D3D \cite{predator2d3d}} & 86.1 & 89.2 & 63.9 & 24.3 & 65.9 \\
\multicolumn{1}{l|}{2D3D-MATR \cite{matr2d3d}} & 
\underline{\textbf{98.6}} & 
\underline{98.0} & 
\underline{88.7} & 
\underline{77.9} & 
\underline{90.8} \\
\multicolumn{1}{l|}{FreeReg \cite{freereg}} & 91.9 & 93.4 & 93.1 & 49.6 & 82.0 \\
\multicolumn{1}{l|}{B2-3Dnet \cite{b23d}} & \textcolor{teal}{\textbf{100.0}} & \textbf{99.0} & \textbf{92.8} & \textcolor{teal}{\textbf{85.8}} & \textcolor{teal}{\textbf{94.4}} \\
\multicolumn{1}{l|}{CA-I2P \cite{cai2p}} & \textcolor{teal}{\textbf{100.0}} & \textcolor{teal}{\textbf{100.0}} & 91.8 & 82.7 & 93.6 \\
\multicolumn{1}{l|}{Flow-I2P \cite{flowi2p}} & \textcolor{teal}{\textbf{100.0}}& \textcolor{teal}{\textbf{100.0}} &  \textcolor{teal}{\textbf{94.5}} & 78.7 & 93.3 \\

\multicolumn{1}{l|}{M23D (ours)} & 
\textcolor{teal}{\textbf{100.0}} & 
\textcolor{teal}{\textbf{100.0}} & 
\textbf{92.8} & 
\textbf{83.6} & 
\textbf{94.1} \\ \midrule
\multicolumn{6}{c}{\textit{Registration Recall} ↑} \\ \midrule
\multicolumn{1}{l|}{FCGF-2D3D \cite{fcgf2d3d}} & 26.5 & 41.2 & 37.1 & 16.8 & 30.4 \\
\multicolumn{1}{l|}{P2-Net \cite{p2}} & 40.3 & 40.2 & 41.2 & 31.9 & 38.4 \\
\multicolumn{1}{l|}{Predator-2D3D \cite{predator2d3d}} & 44.4 & 41.2 & 21.6 & 13.7 & 30.2 \\
\multicolumn{1}{l|}{2D3D-MATR \cite{matr2d3d}} & 
\underline{63.9} & 
\underline{53.9} & 
\underline{58.8} & 
\underline{49.1} & 
\underline{56.4} \\
\multicolumn{1}{l|}{FreeReg+Kabsch \cite{freereg}} & 38.7 & 51.6 & 30.7 & 15.5 & 34.1 \\
\multicolumn{1}{l|}{FreeReg+PnP \cite{freereg}} & 74.2 & 72.5 & 54.5 & 27.9 & 57.3 \\
\multicolumn{1}{l|}{B2-3Dnet \cite{b23d}} & 58.3 & 60.8 & \textbf{74.2} & 60.2 & 63.4 \\ 
\multicolumn{1}{l|}{CA-I2P \cite{cai2p}} & 68.1 & \textbf{73.5} & 63.9 & 47.8 & 63.3 \\
\multicolumn{1}{l|}{Flow-I2P \cite{flowi2p}} & \textcolor{teal}{\textbf{90.0}} & 65.9 & 54.8 & \textbf{63.0} & \textbf{68.4} \\
\multicolumn{1}{l|}{M23D (ours)} & 
\textbf{88.9} & 
\textcolor{teal}{\textbf{76.5}} & 
\textcolor{teal}{\textbf{83.5}} & 
\textcolor{teal}{\textbf{63.3}} & 
\textcolor{teal}{\textbf{78.0}} \\ \bottomrule\bottomrule
\end{tabular}
}
\label{tab:1}
\end{table}


\begin{table}[t]
\caption{Evaluation results on 7Scenes. \textcolor{teal}{Teal} numbers highlight the best, the second best are \textbf{Boldfaced} and the baseline are \underline{underlined}.}
\centering
\vspace{-10pt}
\setlength{\tabcolsep}{1mm} 
\resizebox{\columnwidth}{!}{
\begin{tabular}{@{}l@{\hskip 1pt}|cccccccc@{}}
\toprule\toprule
\multicolumn{1}{l|}{Model} & Chs & Fr & Hds & Off & Pmp & Kit & Strs & Mean \\ \midrule
\multicolumn{1}{l|}{Mean depth(m)} & 1.78 & 1.55 & 0.80 & 2.03 & 2.25 & 2.13 & 1.84 & 1.77 \\ \midrule
\multicolumn{9}{c}{\textit{Inlier Ratio} \(\uparrow\)} \\ \midrule
\multicolumn{1}{l|}{FCGF-2D3D \cite{fcgf2d3d}} & 34.2 & 32.8 & 14.8 & 26 & 23.3 & 22.5 & 6.0 & 22.8 \\
\multicolumn{1}{l|}{P2-Net \cite{p2}} & 55.2 & 46.7 & 13.0 & 36.2 & 32.0 & 32.8 & 5.8 & 31.7 \\
\multicolumn{1}{l|}{Predator-2D3D \cite{predator2d3d}} & 34.7 & 33.8 & 16.6 & 25.9 & 23.1 & 22.2 & 7.5 & 23.4 \\
\multicolumn{1}{l|}{2D3D-MATR \cite{matr2d3d}} & \underline{72.1} & \underline{66.0} & \underline{31.3} & \underline{60.7} & \underline{50.2} & \underline{52.5} & \underline{18.1} & \underline{50.1} \\

\multicolumn{1}{l|}{B2-3Dnet \cite{b23d}} & 73.8 & \textbf{66.7} & 33.1 & 61.7 & 50.8 & 52.3 & 18.1 & 50.9 \\
\multicolumn{1}{l|}{CA-I2P \cite{cai2p}} & 73.6 & 66.4 & 34.5 & \textbf{62.4} & 52.1 & \textbf{52.8} & \textcolor{teal}{\textbf{19.1}} & 51.6 \\
\multicolumn{1}{l|}{Flow-I2P \cite{flowi2p}} & \textcolor{teal}{\textbf{76.6}} & 64.7 & \textbf{37.1} & 62.0 & \textbf{52.3} & \textbf{52.8} & 18.5 & \textbf{52.0} \\

\multicolumn{1}{l|}{M23D(ours)} & 
\textbf{75.0} & 
\textcolor{teal}{\textbf{68.3}} & 
\textcolor{teal}{\textbf{37.7}} & 
\textcolor{teal}{\textbf{65.5}} & 
\textcolor{teal}{\textbf{53.1}} & 
\textcolor{teal}{\textbf{55.2}} & 
\textbf{18.7} & 
\textcolor{teal}{\textbf{53.4}} \\ \midrule
\multicolumn{9}{c}{\textit{Feature Matching Recall} \(\uparrow\)} \\ \midrule
\multicolumn{1}{l|}{FCGF-2D3D \cite{fcgf2d3d}} & \textbf{99.7} & 98.2 & 69.9 & 97.1 & 83.0 & 87.7 & 16.2 & 78.8 \\
\multicolumn{1}{l|}{P2-Net \cite{p2}} & \textcolor{teal}{\textbf{100.0}} & 99.3 & 58.9 & 99.1 & 87.2 & 92.2 & 16.2 & 79 \\
\multicolumn{1}{l|}{Predator-2D3D \cite{predator2d3d}} & 91.3 & 95.1 & 76.6 & 88.6 & 79.2 & 80.6 & 31.1 & 77.5 \\
\multicolumn{1}{l|}{2D3D-MATR \cite{matr2d3d}} & 
\underline{\textcolor{teal}{\textbf{100.0}}} & 
\underline{99.6} & 
\underline{98.6} & 
\underline{\textcolor{teal}{\textbf{100.0}}} & 
\underline{92.4} & 
\underline{95.9} & 
\underline{58.2} & 
\underline{92.1} \\
\multicolumn{1}{l|}{B2-3Dnet \cite{b23d}} & \textcolor{teal}{\textbf{100.0}} & \textcolor{teal}{\textbf{100.0}} & \textbf{98.6} & \textcolor{teal}{\textbf{100.0}} & 92.7 & 95.6 & \textbf{\textcolor{teal}{64.9}} & \textbf{\textcolor{teal}{93.1}} \\ 
\multicolumn{1}{l|}{CA-I2P \cite{cai2p}} & \textcolor{teal}{\textbf{100.0}} & \textbf{\textcolor{teal}{100.0}} & \textbf{98.6} & \textcolor{teal}{\textbf{100.0}} & 92.0 & 95.5 & \textbf{60.8} & 92.4 \\
\multicolumn{1}{l|}{Flow-I2P \cite{flowi2p}} & \textcolor{teal}{\textbf{100.0}} & \textbf{99.7} & 95.1 & \textbf{99.9} & \textbf{\textcolor{teal}{93.1}} & \textbf{\textcolor{teal}{96.8}} & 56.7 & 91.6 \\
\multicolumn{1}{l|}{M23D(ours)} & 
\textcolor{teal}{\textbf{100.0}} & 
\textcolor{teal}{\textbf{100.0}} & 
\textcolor{teal}{\textbf{100.0}} & 
\textcolor{teal}{\textbf{100.0}} & 
\textcolor{teal}{\textbf{93.8}} & 
\textbf{96.0} & 
59.5 & 
\textbf{92.8} \\ \midrule
\multicolumn{9}{c}{\textit{Registration Recall} \(\uparrow\)} \\ \midrule
\multicolumn{1}{l|}{FCGF-2D3D \cite{fcgf2d3d}} & 89.5 & 79.7 & 19.2 & 85.9 & 69.4 & 79.0 & 6.8 & 61.4 \\
\multicolumn{1}{l|}{P2-Net \cite{p2}} & 96.9 & 86.5 & 20.5 & 91.7 & 75.3 & 85.2 & 4.1 & 65.7 \\
\multicolumn{1}{l|}{Predator-2D3D \cite{predator2d3d}} & 69.6 & 60.7 & 17.8 & 62.9 & 56.2 & 62.6 & 9.5 & 48.5 \\
\multicolumn{1}{l|}{2D3D-MATR \cite{matr2d3d}} & 
\underline{96.9} & 
\underline{90.7} & 
\underline{52.1} & 
\underline{95.5} & 
\underline{80.9} & 
\underline{86.1} & 
\underline{28.4} & 
\underline{75.8} \\
\multicolumn{1}{l|}{B2-3Dnet \cite{b23d}} & 98.3 & 90.5 & 56.2 & \textbf{96.4} & \textbf{84.0} & 86.1 & 32.4 & 77.7 \\ 
\multicolumn{1}{l|}{CA-I2P \cite{cai2p}} & \textbf{\textcolor{teal}{99.0}} & \textbf{90.7} & \textcolor{teal}{\textbf{68.5}} & 96.2 & 83.0 & 88.1 & 31.1 & \textbf{79.5} \\
\multicolumn{1}{l|}{Flow-I2P \cite{flowi2p}} & \textbf{98.8} & 90.0 & 58.4 & 93.9 & 82.1 & \textbf{88.6} & \textbf{\textcolor{teal}{37.6}} & 78.4 \\
\multicolumn{1}{l|}{M23D(ours)} & 
97.6 & 
\textcolor{teal}{\textbf{95.6}} & 
\textbf{67.1} & 
\textcolor{teal}{\textbf{98.7}} & 
\textcolor{teal}{\textbf{84.4}} & 
\textcolor{teal}{\textbf{89.0}} & 
\textbf{35.1} & 
\textcolor{teal}{\textbf{81.3}} \\ \bottomrule\bottomrule
\end{tabular}
}
\vspace{-10pt}
\label{tab:2}
\end{table}


\textbf{Metrics.} We evaluate our method using several key metrics:  Inlier Ratio (IR), Feature Matching Recall (FMR), and Registration Recall (RR).

\textit{Inlier Ratio} (IR) quantifies the proportion of inliers among all putative pixel-point correspondences. A correspondence is deemed an inlier if its 3D distance is less than a threshold \(\tau_1 = 5 \text{ cm}\) under the ground-truth transformation \(\mathbf{T}^*_{\mathcal{P} \rightarrow \mathcal{I}}\):
\begin{equation}
\text{IR} = \frac{1}{|C|} \sum_{(x_i, y_i) \in C} \left[ \left\| \mathbf{T}^*_{\mathcal{P} \rightarrow \mathcal{I}}(x_i) - \mathbf{K}^{-1}(y_i) \right\|_2 < \tau_1 \right].
\end{equation}

Here, \([ \cdot ]\) denotes the Iverson bracket, \(x_i \in \mathcal{P}\), and \(y_i \in \mathcal{Q} \subseteq \mathcal{I}\) are pixel coordinates. The function \(\mathbf{K}^{-1}\) projects a pixel to a 3D point based on its depth value.

 \textit{Feature Matching Recall} (FMR) represents the fraction of image-point-cloud pairs with an IR above a threshold \(\tau_2 = 0.1\). It measures the likelihood of successful registration:
\begin{equation}
\text{FMR} = \frac{1}{M} \sum_{i=1}^{M} \left[ \text{IR}_i > \tau_2 \right],
\end{equation}
where \(M\) is the total number of image-point-cloud pairs.

 \textit{Registration Recall} (RR) measures the fraction of image-point-cloud pairs that are correctly registered. A pair is correctly registered if the root mean square error (RMSE) between the ground-truth-transformed and predicted point clouds \(\mathbf{T}_{\mathcal{P} \rightarrow \mathcal{I}}\) is less than \(\tau_3 = 0.1\text{ m}\):
\begin{equation}
\text{RMSE} = \sqrt{\frac{1}{|\mathcal{P}|} \sum_{p_i \in \mathcal{P}} \left\| \mathbf{T}_{\mathcal{P} \rightarrow \mathcal{I}}(p_i) - \mathbf{T}^*_{\mathcal{P} \rightarrow \mathcal{I}}(p_i) \right\|_2^2},
\end{equation}
\begin{equation}
\text{RR} = \frac{1}{M} \sum_{i=1}^{M} \left[ \text{RMSE}_i < \tau_3 \right].
\end{equation}

\begin{table}[!t]
\small
\centering
\caption{Evaluation Results on KITTI Dataset. \textcolor{teal}{Teal} numbers highlight the best.}
\resizebox{\columnwidth}{!}{
\begin{tabular}{l|l|c|c}
\toprule\toprule
\textbf{Method} & \textbf{Type} & \textbf{RTE(m) $\downarrow$} & \textbf{RRE ($^\circ$) $\downarrow$} \\
\midrule
vpc + GeoTransformer\cite{geotransformer} & Point-to-Point & 4.27 $\pm$ 7.14 & 8.67 $\pm$ 8.55 \\
vpc + Hunter & Point-to-Point & 4.59 $\pm$ 5.22 & 6.23 $\pm$ 5.17 \\
DeepI2P(2D) \cite{deepi2p} & Image-to-Point & 5.15 $\pm$ 7.35 & 9.14 $\pm$ 8.02 \\
CorrI2P \cite{corri2p} & Image-to-Point & 4.24 $\pm$ 7.26 & 6.47 $\pm$ 5.20 \\
VP2P \cite{vp2p} & Image-to-Point & 2.05 $\pm$ 3.23 & 4.01 $\pm$ 6.37 \\
2D3D-MATR \cite{matr2d3d} & Image-to-Point & 1.86 $\pm$ 3.79 & 2.59 $\pm$ 4.46 \\
RetrI2P \cite{retri2p} & Image-to-Point & 1.61 $\pm$ 2.39 & 3.16 $\pm$ 2.85 \\
FreeReg \cite{freereg} & Image-to-Point & 1.78 $\pm$ 1.76 & 2.89 $\pm$ 4.47 \\
CFI2P \cite{cfi2p} & Image-to-Point & 1.95 $\pm$ 2.97 & 2.63 $\pm$ 3.19 \\
M23D (Ours) & Image-to-Point & \textcolor{teal}{\textbf{1.58 $\pm$ 2.33}} & \textcolor{teal}{\textbf{2.34 $\pm$ 3.29}} \\
\bottomrule\bottomrule
\end{tabular}
}
\label{tab:kitti}
\end{table}

\begin{table}[!b]
\vspace{-15pt}
\caption{Ablation studies on 7Scenes. \textcolor{teal}{Teal} numbers highlight the best.}
\centering
\setlength{\tabcolsep}{1mm}
\resizebox{\columnwidth}{!}{
\begin{tabular}{@{}cl|c|c|c|ccc@{}}
\toprule
\multicolumn{2}{l|}{Model} & \multicolumn{1}{l|}{Image-MAE} & Point-MAE & Cross Modal & IR  & FMR  & RR   \\ \midrule
\multicolumn{2}{c|}{BL}   &         &       &       & 50.1 & 92.1 & 75.8 \\ \midrule
\multicolumn{2}{c|}{F1}   & \checkmark &       &       & 49.8 & 91.2 & 78.0 \\ \midrule
\multicolumn{2}{c|}{F2}   &         & \checkmark &       & 52.2 & 92.4 & 78.3 \\ \midrule
\multicolumn{2}{c|}{F3}   & \checkmark & \checkmark &       & 50.4 & 92.0 & 80.4 \\ \midrule
\multicolumn{2}{c|}{Full} & \checkmark & \checkmark & \checkmark &
\textcolor{teal}{\textbf{53.4}} &
\textcolor{teal}{\textbf{92.8}} &
\textcolor{teal}{\textbf{81.3}} \\ \bottomrule
\end{tabular}
}
\label{tab3}
\end{table}

These metrics provide a comprehensive evaluation of the model's capability to accurately match features and register image-point-cloud pairs, ensuring robust alignment and effective correspondence.

\subsection{Evaluations on Dataset}

We compare our approach with 2D3D-MATR \cite{matr2d3d} and other baseline methods \cite{fcgf2d3d,p2,predator2d3d,freereg,b23d,cai2p,flowi2p} on the RGB-D Scenes v2 dataset, as shown in Table~\ref{tab:1}.
Our method integrates the ID-MAE framework to enhance feature extraction and better align the image and point cloud modalities. The introduction of the SRLM strategy further refines the model by adaptively focusing on semantically meaningful regions while suppressing noise from irrelevant areas. These improvements lead to a substantial increase in performance across all evaluation metrics. Specifically, our method achieves a 11.1 percentage point gain in inlier ratio (from 32.4 to 43.5), a 3.3 percentage point improvement in feature matching recall (from 90.8 to 94.1), and a 21.6 percentage point increase in registration recall (from 56.4 to 78.0) compared to the baseline method. Notably, our method sets a new state-of-the-art in registration recall, with a remarkable improvement of 7.8 percentage points, demonstrating its robustness in accurately estimating correspondences across different modalities.

\begin{figure*}[!t]
\centering
\includegraphics[width=0.9\textwidth]{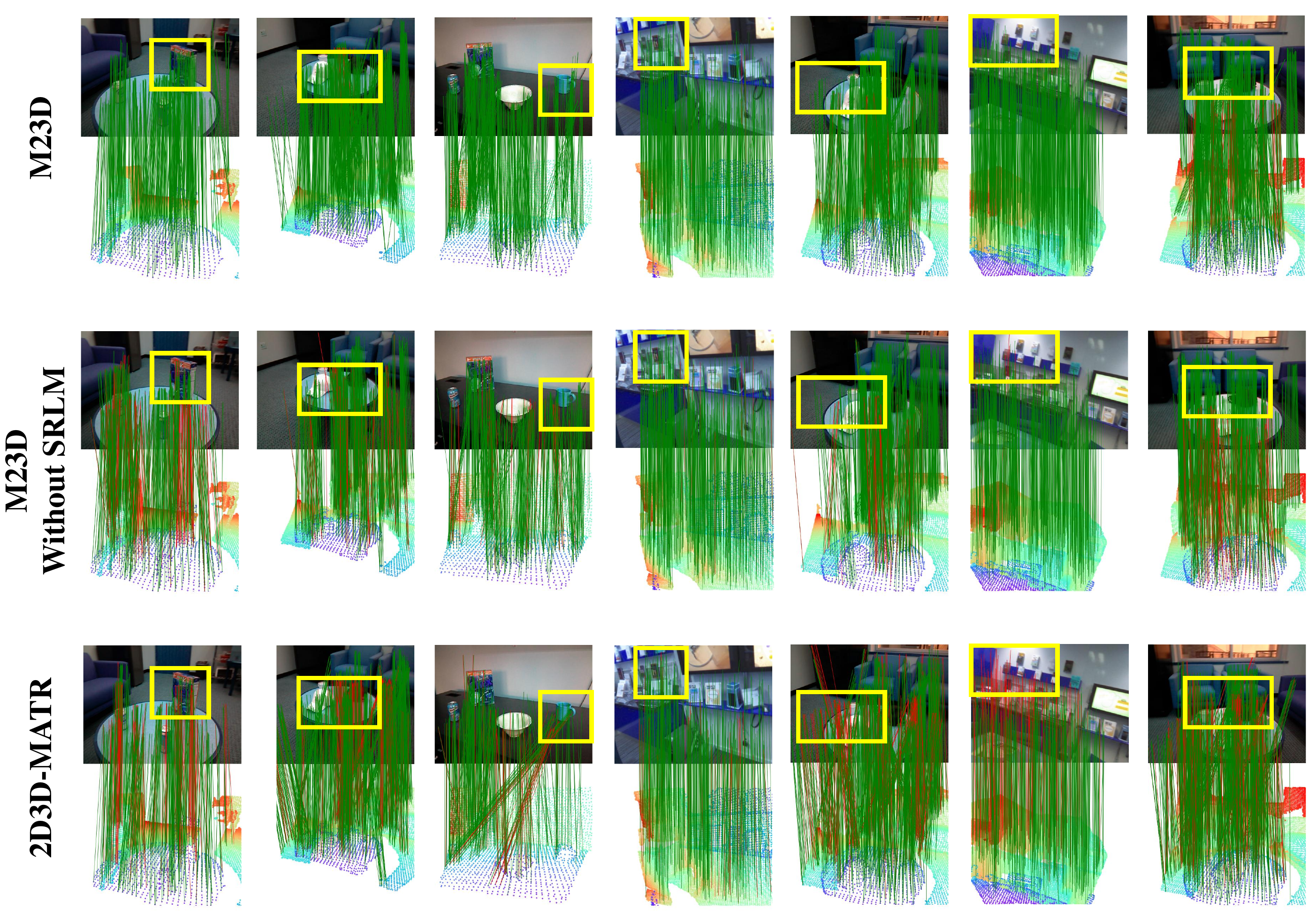}
\vspace{-15pt}
\caption{The visualization results of M23D. To rigorously analyze the performance, we set the error threshold to a strict 50px.}
\label{vis}
\vspace{-15pt}
\end{figure*}

\begin{table}[!b]
\vspace{-15pt}
\caption{Ablation studies of different masking strategies on RGB-D Scenes v2. \textbf{\textcolor{teal}{Teal}} numbers highlight the best.}
\centering
\setlength{\tabcolsep}{1mm} 
\renewcommand{\arraystretch}{1.5} 
\resizebox{\columnwidth}{!}{
\begin{tabular}{c||cccc||cccc}
\toprule
Method & Random & Similarity & RL & SRLM & GFLOPs & IR & FMR & RR \\
\midrule
M1 &   &   &   &   & 388.9 & 32.5 & 91.0 & 56.4 \\
M2 &  \checkmark  &  &   &   & 423.8 & 42.9 & 93.2 & 63.5 \\
M3 &  & \checkmark  &   &   & 457.2 & 37.1 & 92.9 & 67.2 \\
M4 &   &   & \checkmark &   & 834.8 & 42.7 & 93.4 & 73.1 \\
M5 &  &  &  & \checkmark & \textbf{\textcolor{teal}{535.1}} & \textbf{\textcolor{teal}{43.5}} & \textbf{\textcolor{teal}{94.1}} & \textbf{\textcolor{teal}{78.0}} \\
\bottomrule
\end{tabular}}
\label{tab:ablation}
\end{table}

Compared to RGB-D Scenes v2, 7-Scenes exhibit greater scale variations, yet our method still outperforms others, as shown in Table~\ref{tab:2}. In the Inlier Ratio metric, we achieve 53.4\%, surpassing 2D3D-MATR by 3.3 percentage points (50.1\%). For Feature Matching Recall, our method reaches 92.8\%, demonstrating robustness in correspondence establishment, outperforming 2D3D-MATR by 0.7 percentage points (92.1\%). In Registration Recall, we achieve 81.3\%, maintaining an advantage of 5.5 percentage points over 2D3D-MATR's 75.8\%. Notably, 2D3D-MATR shows improvements in the challenging Heads and Pumpkin scenes, with the Heads scene involving small errors due to texture-less areas, and the Pumpkin scene containing repetitive patterns. Our ID-MAE effectively addresses these challenges, ensuring superior performance across all scenes.
Although B2-3Dnet achieves slightly better performance in FMR due to its targeted modeling of image features, our method offers a greater overall advantage when considering the comprehensive results.

\begin{figure}[!b]
\centering
\vspace{-10pt}
\includegraphics[width=\columnwidth]{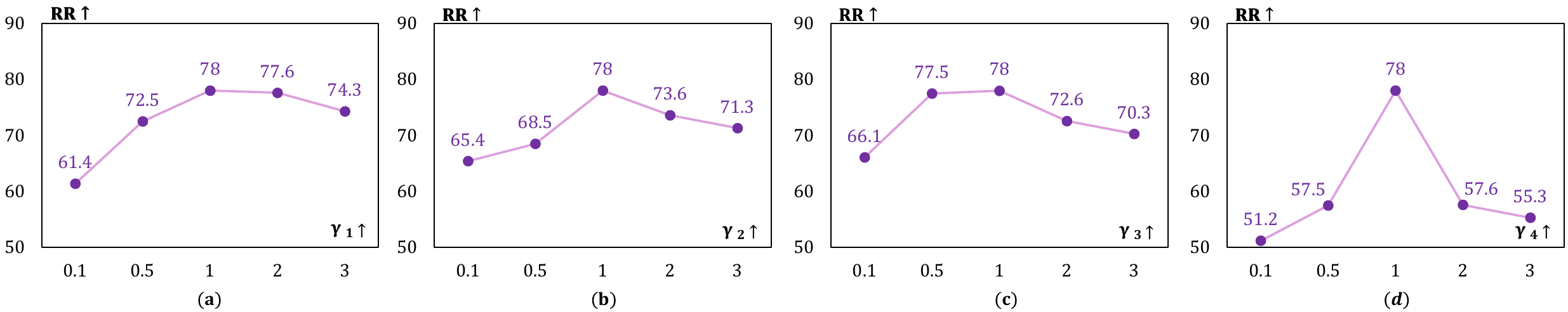}
\vspace{-18pt}
\caption{Ablation studies on the hyperparameters.}
\label{ablation}
\end{figure}

To better evaluate our performance, we conducted experiments on the outdoor dataset KITTI to verify the robustness of our method.
As shown in Table~\ref{tab:kitti}, we evaluate our method, M23D, on the KITTI dataset and compare it with several state-of-the-art techniques. Our evaluation is based on two key metrics: Reprojection Error (RTE) and Rotation Error (RRE), both of which are essential for assessing the accuracy of 3D point cloud registration. Our method outperforms all other approaches in both RTE and RRE. Specifically, M23D achieves a significantly lower RTE of 1.58 $\pm$ 2.33 meters and a lower RRE of 2.34 $\pm$ 3.29 degrees. This surpasses other image-to-point methods, such as RetrI2P (RTE: 1.61 $\pm$ 2.39 meters, RRE: 3.16 $\pm$ 2.85 degrees) and VP2P (RTE: 2.05 $\pm$ 3.23 meters, RRE: 4.01 $\pm$ 6.37 degrees).
These results demonstrate that our method provides more accurate alignment with lower error margins compared to previous works. The results on the KITTI dataset confirm the robustness and superiority of M23D over existing methods, highlighting its potential for real-world applications that require high-precision registration.

\subsection{Ablation Studies}

We perform ablation studies on the 7Scenes dataset to evaluate the contributions of different components in our model, as shown in Table~\ref{tab3}. Compared to the baseline (BL), adding Image-MAE (F1) or Point-MAE (F2) slightly improves the performance. The Dual MAE (F3) leads to further improvement. After introducing Cross-Modal alignment (Full), the model achieves the best results, demonstrating the effectiveness of our full approach.

\begin{figure*}[!t]
\centering
\includegraphics[width=0.9\textwidth]{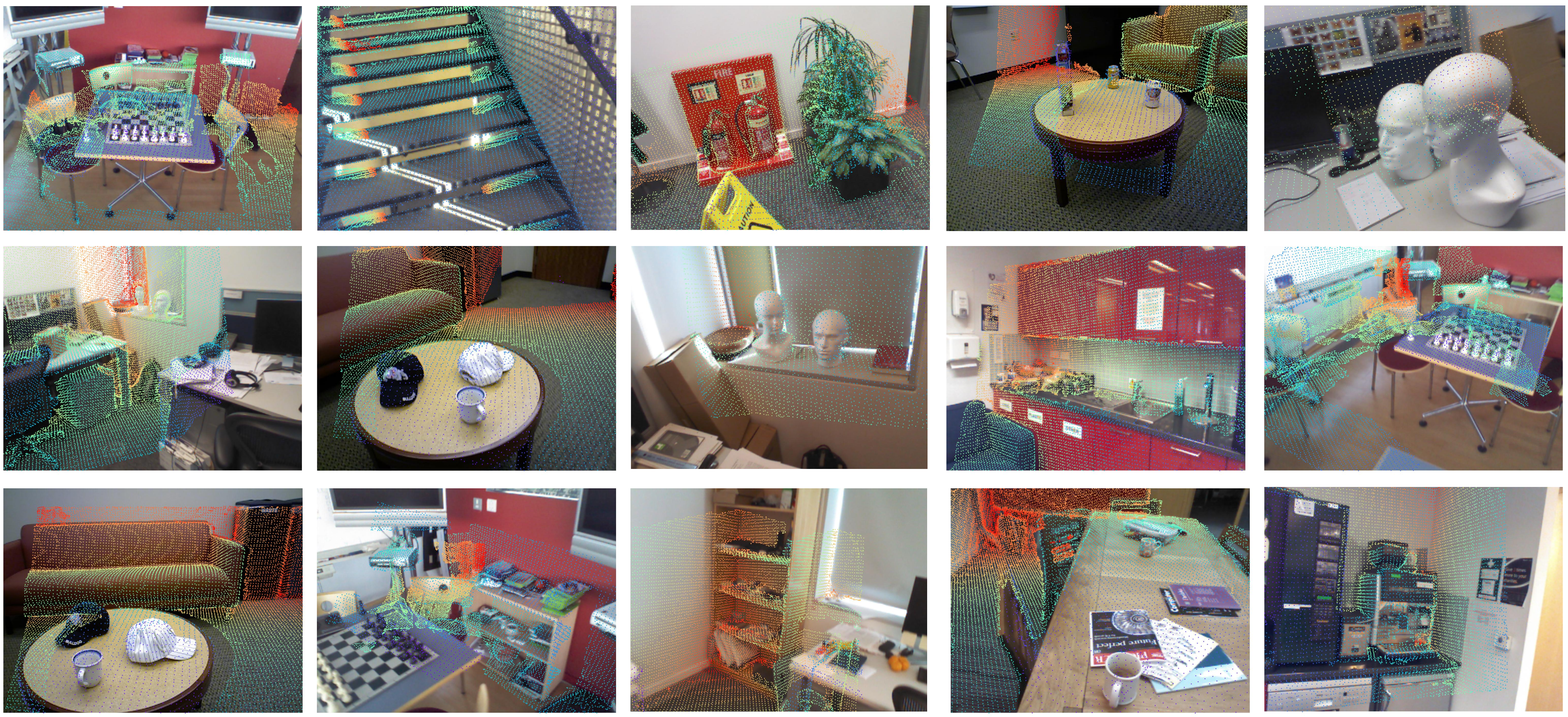}
\vspace{-5pt}
\caption{The visualization of the point cloud projection onto the image demonstrates that our method achieves precise rigid transformation, maintaining alignment without significant discrepancies.}
\label{vis2}
\vspace{-12pt}
\end{figure*}

We conduct ablation studies to evaluate the impact of different masking strategies on the RGB-D Scenes v2 dataset, as shown in Table~\ref{tab:ablation}. Compared to the baseline, random masking (M2) improves performance; however, due to its lack of targeted selection, the gains remain limited. Similarity-based masking (M3) further improves the results with only a slight increase in computational cost. Although directly applying reinforcement learning (M4) achieves strong performance, it incurs a significant increase in computation. In contrast, our SRLM strategy (M5) achieves the highest performance while maintaining good efficiency, demonstrating the superiority of our design in balancing accuracy and computational cost.
{Our strategy is specifically designed to adapt to the MAE framework. Therefore, the effectiveness of the reinforcement learning (RL) component serves as evidence of the strong capability of our MAE-based design. Previous common masking strategies often limit the full potential of the MAE module, whereas our approach enables more effective utilization of its representation learning capacity.
The inter-modal MAE serves as the core representation learning mechanism, enforcing cross-modal consistency through reconstruction and establishing a strong feature foundation for 2D–3D correspondence estimation. The RL module is introduced solely to optimize the mask selection process, addressing the non-differentiability of discrete masking and improving training efficiency, rather than replacing or weakening the role of MAE.
Importantly, MAE-based supervision alone already brings consistent performance gains compared to the baseline. The RL strategy further refines the masking process by prioritizing informative regions, leading to additional improvements on top of the MAE framework.
While similarity evaluation is widely used in registration tasks, it is typically applied at the correspondence or matching stage. In contrast, our similarity-aware masking strategy leverages these cues during the representation learning phase. By guiding the MAE to focus on informative cross-modal regions before correspondence estimation, we shift the usage of similarity. Rather than directly establishing correspondences, similarity is used to prioritize which regions should be preserved or reconstructed during the masked modeling process. This allows the MAE to learn more correspondence-aware representations early in the process, setting the stage for more effective matching and correspondence estimation later.}

{
\begin{table}[!b]
\vspace{-10pt}
\centering
\caption{Efficiency comparison of different methods.}
\label{tab:efficiency}
\setlength{\tabcolsep}{6pt}
\renewcommand{\arraystretch}{1.05}
\resizebox{0.9\columnwidth}{!}{%
\begin{tabular}{lccc}
\toprule
Model & 2D3D-MATR & Flow-I2P & Ours \\
\midrule
FPS $\uparrow$        & 7.852 & 6.061 & 7.684 \\
GFLOPs $\downarrow$   & 388.2 & 525.1 & 535.1 \\
Memory (GB) $\downarrow$ & 5.637 & 2.844 & 5.423 \\
Param (M) $\downarrow$   & 28.2 & 28.7 & 28.6 \\
\bottomrule
\end{tabular}%
}
\end{table}

Fig.~7 reports ablation studies on the key hyperparameters in our framework, where the registration recall (RR) is evaluated under different parameter settings. As shown in Fig.~7(a)--(c), the performance consistently improves when increasing the corresponding hyperparameters from small values, and reaches a relatively stable region around the default setting (e.g., $\gamma_1=\gamma_2=\gamma_3=1$). Further increasing these values leads to marginal performance fluctuations or slight degradation, indicating that overly strong weighting may introduce redundancy or reduce robustness.  
Fig.~7(d) exhibits a similar trend, where moderate values yield the best performance, while extreme settings result in noticeable drops. Overall, these results demonstrate that the proposed method is not overly sensitive to hyperparameter choices and achieves stable performance within a reasonable range, validating the robustness of our design and the practicality of the selected default parameters.

Table~VI compares the efficiency of different detection-free image-to-point cloud registration methods in terms of inference speed, computational cost, memory usage, and model size. Our method achieves an inference speed of 7.684 FPS, which is comparable to 2D3D-MATR (7.852 FPS) and faster than Flow-I2P (6.061 FPS), indicating that no additional inference latency is introduced. Although our approach requires slightly higher computational cost (535.1 GFLOPs) than 2D3D-MATR, this overhead mainly arises from similarity computation and does not affect inference efficiency, since the proposed intermodal MAE and RL-based masking strategy are applied only during training. Moreover, our method maintains a comparable memory footprint and parameter count (28.6M), demonstrating that the improved registration performance is achieved without significantly increasing model complexity.
Compared to our method, Flow-I2P exhibits better performance in terms of GFLOPs and memory usage, but lags slightly in FPS. 
Overall, the results suggest that our approach strikes a favorable balance between accuracy and efficiency, making it suitable for practical deployment.}

\begin{figure}[!t]
\centering
\includegraphics[width=\columnwidth]{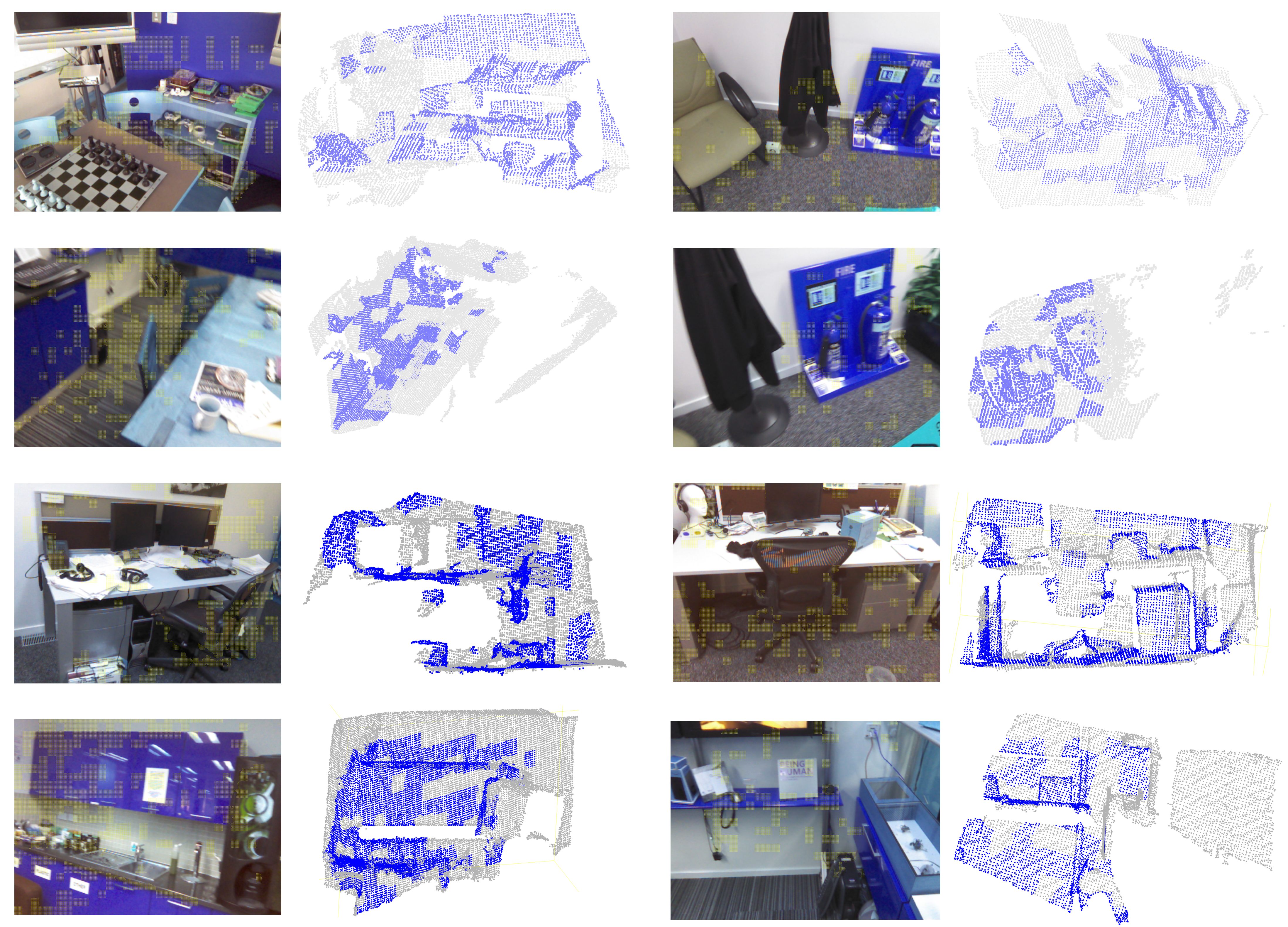}
\vspace{-15pt}
\caption{Visualization of image and point cloud corresponding regions selected by reinforcement learning. The selected regions are highlighted in yellow in the image and in blue in the point cloud.}
\label{region}
\vspace{-18pt}
\end{figure}

\subsection{Visualization}

We conduct extensive experiments to validate the effectiveness of our model and visualize M23D’s matching performance for a more intuitive understanding. As shown in Fig.~\ref{vis}, we select seven scene pairs from the 7Scenes and RGB-D Scenes v2 datasets.  
In each visualization, the top, middle, and bottom rows correspond to M23D, M23D without SRLM, and the baseline model, respectively. We consider matches with a projection distance below 50 pixels as correct (green lines), and others as incorrect (red lines), using this threshold for visualization.
The results show that applying ID-MAE improves matching, with more green lines indicating better correspondence. However, random masking still leads to some mismatches (red lines). With SRLM, matching performance improves significantly, demonstrating the effectiveness of our approach in enhancing both accuracy and robustness for image-to-point cloud registration.
In Fig.~\ref{vis2}, we project the point cloud onto the image using the obtained rigid transformation. It can be seen that the structure aligns well, with no large-scale misalignments. Especially in the Heads scene, which lacks sufficient texture information, the registration results are still maintained effectively.

{The visualization in Fig. 9 shows the image and point cloud regions selected by the proposed reinforcement learning-based masking strategy during training. The highlighted yellow regions in the image and blue regions in the point cloud indicate the areas considered informative for cross-modal matching.
It can be observed that these selected regions tend to concentrate on areas with high texture and prominent structures. This is likely because these regions can provide high-quality features, and applying the Masked Auto-Encoder (MAE) to these regions can significantly improve their representation learning, thereby enhancing the overall registration accuracy.
The selective masking strategy focuses the model's attention on the most informative parts of the input data, allowing it to learn robust cross-modal correspondences from these salient regions. By emphasizing the feature-rich areas, the model can better capture the intricate geometric and visual cues needed for accurate image-to-point cloud registration, leading to the improved performance demonstrated by the method.
This approach is a key innovation of the proposed technique, as it enables efficient and effective learning of the cross-modal mapping without requiring the model to process the entire input indiscriminately. The targeted focus on the most discriminative regions contributes to the overall efficiency and accuracy of the registration pipeline. It can be observed that the selected regions in the image and the point cloud do not always exhibit a strict one-to-one correspondence. This behavior is expected, as the masking strategy operates at the feature and region level, aiming to preserve discriminative semantic and geometric regions rather than enforcing precise pixel-to-point alignment. Moreover, differences in viewpoint, sensing modality, and partial visibility naturally lead to asymmetric region selection between the image and point cloud domains.

\begin{table}[!t]
\centering
\caption{Generalization ability of SRLM across different datasets.}
\label{tab:gen}
\setlength{\tabcolsep}{6pt}
\renewcommand{\arraystretch}{1.05}
\resizebox{0.9\columnwidth}{!}{
\begin{tabular}{@{}lllll@{}}
\toprule
Training Dataset & Testing Dataset & Mask Strategy & IR & RR \\
\midrule
7-Scenes & RGB-D Scenes v2 & Random Mask & 18.1 & 23.9 \\
\midrule
7-Scenes & RGB-D Scenes v2 & SRLM & 18.7 & 24.3 \\
\midrule\midrule
7-Scenes & ScanNet & Random Mask & 15.9 & 24.1 \\
\midrule
7-Scenes & ScanNet & SRLM & 16.2 & 28.7 \\
\bottomrule
\end{tabular}}
\vspace{-18pt}
\end{table}

These visualizations also reveal several limitations of the proposed method. First, similar to most learning-based image-to-point cloud registration approaches, in low-overlap scenarios (e.g., the bottom-right example in Fig.~9), reliable cross-modal cues are inherently scarce, and the selected regions tend to concentrate on limited areas, which weakens the overall discriminative capability of the masking strategy. This phenomenon stems from the intrinsic ambiguity of correspondence supervision under minimal overlap and remains a common challenge in the field. Second, the proposed similarity-aware masking strategy relies on an initial cross-modal similarity map to guide informative region selection. While this design improves training stability and efficiency, its effectiveness may degrade when the initial similarity estimation is severely affected by noise or large viewpoint variations. We view this dependency as a reasonable trade-off, and consider improving the robustness of initial similarity estimation as a promising direction for future work. Finally, regarding scalability to large-scale scenes, the current implementation follows standard detection-free registration pipelines and is primarily evaluated on indoor and moderate-scale outdoor datasets. Although the proposed MAE and RL-based masking strategy are applied only during training and introduce no additional inference overhead, extending the framework to larger-scale environments may require hierarchical processing or more memory-efficient similarity computation strategies.

To further evaluate the cross-dataset generalization ability of SRLM, we train the model on 7-Scenes and directly evaluate it on RGB-D Scenes v2 and ScanNet (see in Table~VII). Compared with the random masking strategy, SRLM consistently improves both the Inlier Ratio (IR) and Registration Recall (RR) across the two target datasets. Specifically, on RGB-D Scenes v2, SRLM increases IR from 18.1\% to 18.7\% and RR from 23.9\% to 24.3\%. On ScanNet, SRLM improves IR from 15.9\% to 16.2\%, while achieving a more substantial RR improvement from 24.1\% to 28.7\%, corresponding to a gain of 4.6 percentage points. These consistent improvements indicate that SRLM does not merely overfit the scene distribution of the training dataset. Instead, similarity-based initialization helps suppress irrelevant and non-overlapping regions, while RL-driven refinement adaptively identifies registration-informative regions according to the task-aligned matching reward. Notably, the larger improvement in RR than in IR on ScanNet suggests that SRLM enhances not only the proportion of correct correspondences, but also their geometric reliability for pose estimation. Overall, these results demonstrate that SRLM learns a more transferable masking policy and improves registration robustness under considerable cross-dataset domain shifts.

Overall, these analyses demonstrate both the effectiveness and the current limitations of the proposed approach in cross-modal region selection and correspondence modeling. Future work will focus on more robust cross-modal cue modeling under low-overlap conditions, enhancing the robustness of initial similarity estimation, and developing hierarchical and efficient registration frameworks for large-scale scenes.
}

\section{Conclusion}

We propose M23D, an Adaptive Dual-Masked Autoencoder Network for Image-to-Point Cloud Registration. Our method introduces a dual MAE to enhance feature extraction by aligning features and establishing correspondences. Additionally, we propose a similarity-based RL masking strategy that reduces unnecessary masking and focuses on key matching regions, improving registration accuracy and robustness. M23D achieves state-of-the-art performance on the RGB-D Scenes v2 and 7-Scenes datasets.

\section*{Acknowledgment}
This work was supported in part by the Fundamental Research Funds for the Central Universities (11040-03742026007).

\begin{IEEEbiography}[{\includegraphics[width=1in,height=1.25in,clip,keepaspectratio]{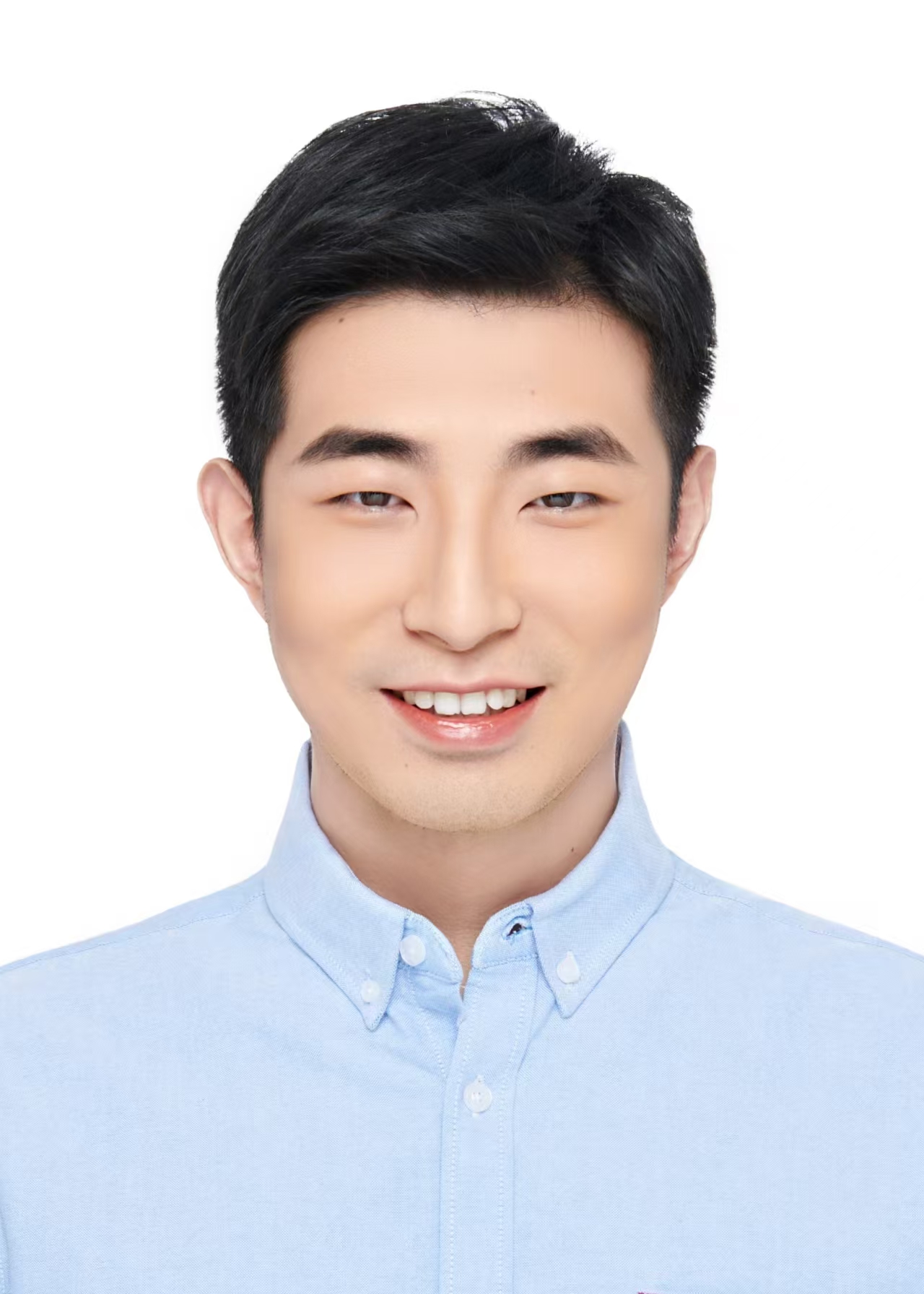}}]{Zhixin Cheng}
 received his bachelor's degree from the School of Electrical and Electronic Engineering, Huazhong University of Science and Technology, Wuhan, China, in 2020, and his Ph.D. degree from the School of Information Science and Technology, University of Science and Technology of China, Hefei, China, through a combined M.S.–Ph.D. program.
He is currently a Specially Appointed Associate Professor with Hefei University of Technology, Hefei, China. His research interests include computer vision and machine learning, with a focus on 3D scene registration, multimodal learning, and image enhancement.
\end{IEEEbiography}

\begin{IEEEbiography}[{\includegraphics[width=1in,height=1.25in,clip,keepaspectratio]{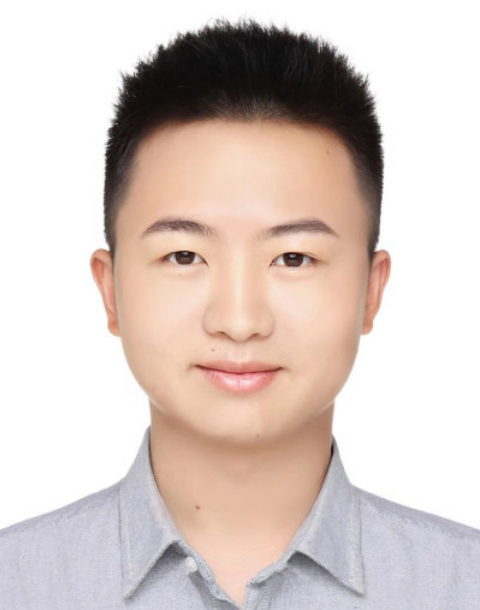}}]{Jiacheng Deng}
  received the bachelor's degree in Information Security from the University of Science and Technology of China in 2023. He is now pursuing a master degree in Control Science and Engineering at University of Science and Technology of China. His research interests include computer vision and deep learning, especially image-to-point cloud registration and pose estimation. 
\end{IEEEbiography}

\begin{IEEEbiography}[{\includegraphics[width=1in,height=1.25in,clip,keepaspectratio]{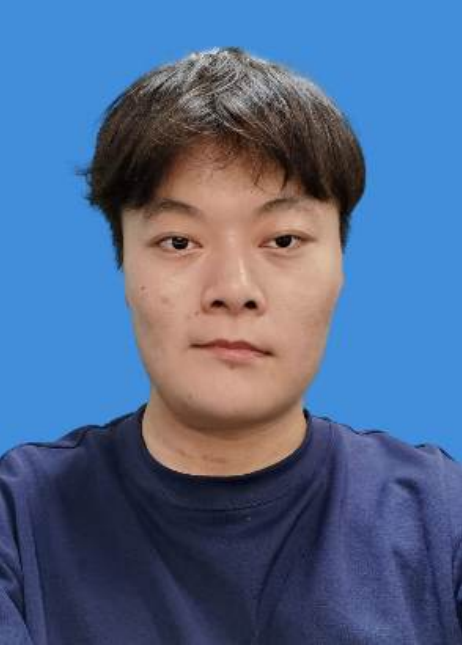}}]{Xiaotian Yin}
    is currently pursuing a Ph.D. degree at the University of Science and Technology of China, Hefei, China. His research interests include computer vision and machine learning, with a focus on few-shot learning and multi-modal learning.
\end{IEEEbiography}

\begin{IEEEbiography}[{\includegraphics[width=1in,height=1.25in,clip,keepaspectratio]{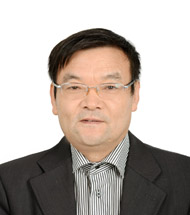}}]{Baoqun Yin}
   received his bachelor’s degree in Mathematics from Sichuan University, Chengdu, China, in 1985, and his master’s degree in Applied Mathematics from the University of Science and Technology of China (USTC), Hefei, China, in 1993. He earned his Ph.D. degree in Pattern Recognition and Intelligent Systems from the Department of Automation, USTC, in 1998. He is currently a Professor in the Department of Automation at the University of Science and Technology of China. His research interests include stochastic systems, system optimization, and information networks, focusing on Markov decision processes, network optimization, and smart energy management.
\end{IEEEbiography}

\begin{IEEEbiography}
[{\includegraphics[width=1in,height=1.25in,clip,keepaspectratio]
{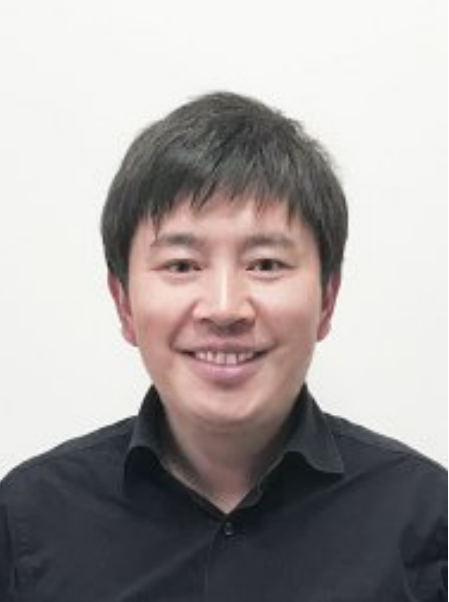}}]
{Richang Hong}
(Senior Member, IEEE) received the Ph.D. degree from the University of
Science and Technology of China, Hefei, China, in 2008. From September
2008 to December 2010, he was a Research Fellow with the School of
Computing, National University of Singapore, Singapore. He is currently
a Professor with the Hefei University of Technology, Hefei, China. He
has coauthored more than 300 publications in the areas of his research
interests, which include multimedia question answering, video content
analysis, and pattern recognition. Dr. Hong is a Member of the
Association for Computing Machinery. He was the recipient of the Best
Paper Award in the ACM Multimedia 2010.
\end{IEEEbiography}

\begin{IEEEbiography}[{\includegraphics[width=1in,height=1.25in,clip,keepaspectratio]{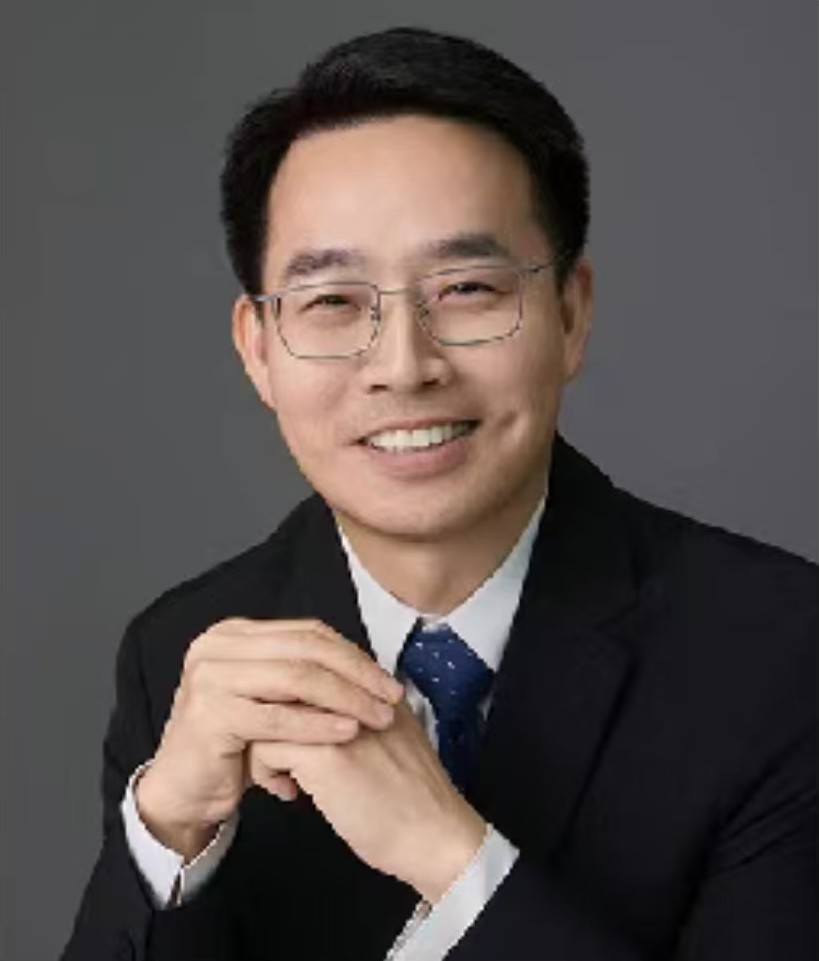}}]{Tianzhu Zhang}
  (M’11) received the bachelor’s degree in Communications and Information Technology from the Beijing Institute of Technology, Beijing, China, in 2006, and the Ph.D. degree in Pattern Recognition and Intelligent Systems from the Institute of Automation, Chinese Academy of Sciences, Beijing, China, in 2011. He is currently a Professor at the Department of Automation, School of Information Science, University of Science and Technology of China. His current research interests include computer vision and multimedia, with a focus on action recognition, object classification, object tracking, and social event analysis.
\end{IEEEbiography}

\vfill

\end{document}